\documentclass[lettersize,journal]{IEEEtran}
\usepackage{amsmath,amsfonts}
\usepackage{algorithmic}
\usepackage{algorithm}
\usepackage{array}
\usepackage{textcomp}
\usepackage{stfloats}
\usepackage{url}
\usepackage{verbatim}
\usepackage{graphicx}
\usepackage{cite}
\usepackage{mathtools} % amsmath with fixes and additions
\usepackage{booktabs} % commands to create good-looking tables
\usepackage{tikz} % nice language for creating drawings and diagrams
\usepackage{subcaption}
\usepackage{bm}
\usepackage{amsthm}
\usepackage{amssymb,physics}
\usepackage{dcolumn}
\newcolumntype{C}{>{\centering}p{1cm}}
\newcolumntype{E}{D{.}{.}{5.0}}
\newtheorem{thm}{Theorem}% theorem counter resets every \subsection
\newtheorem*{theorem*}{Theorem}
\newtheorem*{lemma*}{Lemma}
\newcommand{\Note}[1]{{\textcolor{red} {#1}}}
\newcommand{\blue}[1]{\textcolor{black}{#1}}
\newcommand{\green}[1]{\textcolor{black}{#1}}
\newtheorem{assumption}{Assumption}  
\newtheorem{lemma}{Lemma}

\DeclareMathOperator*{\argmax}{arg\,max}
\DeclareMathOperator*{\argmin}{arg\,min}
\newcommand\numeq[1]%
  {\stackrel{\scriptscriptstyle(\mkern-1.5mu#1\mkern-1.5mu)}{=}}
  \newcommand\numgeq[1]%
  {\stackrel{\scriptscriptstyle(\mkern-1.5mu#1\mkern-1.5mu)}{\geq}}
  \newcommand\numleq[1]%
  {\stackrel{\scriptscriptstyle(\mkern-1.5mu#1\mkern-1.5mu)}{\leq}}
\begin{document}

\title{Robust PAC$^m$: Training Ensemble Models Under Misspecification and Outliers
\thanks{The work of M. Zecchin is funded by the Marie Curie action WINDMILL (Grant agreement No. 813999), while O. Simeone and S. Park have received funding from the European Research Council (ERC) under the European Union’s Horizon 2020 Research and Innovation Programme (Grant Agreement No. 725731). The work of M. Zecchin and O. Simeone was also supported by the European Union’s Horizon Europe project CENTRIC (Grant agreement No. 101096379). The work of O. Simeone has also been supported by an Open Fellowship of the EPSRC with reference EP/W024101/1.  M. Kountouris has received funding from the European Research Council (ERC) under the European Union’s Horizon 2020 Research and Innovation programme (Grant agreement No. 101003431). The work of D. Gesbert was supported by the 3IA artificial intelligence interdisciplinary project French funded by ANR, No. ANR-19-P3IA-0002.

 Marios Kountouris and David Gesbert are with the Communication Systems Department, EURECOM, Sophia-Antipolis, France (e-mail: kountour@eurecom.fr,gesbert@eurecom.fr).

Matteo Zecchin, Sangwoo Park and Osvaldo Simeone are with the King's Communications, Learning \& Information Processing (KCLIP) lab, Department of Engineering, King’s College London, London WC2R 2LS, U.K. (e-mail: matteo.1.zecchin@kcl.ac.uk, sangwoo.park@kcl.ac.uk; osvaldo.simeone@kcl.ac.uk).

The second author has contributed to the definition of technical tools and to the experiments. The third author has had an active role in defining problem and tools, as well as in writing the text, while the last two authors have had a supervisory role.}
}

\author{Matteo Zecchin,~\IEEEmembership{Student Member,~IEEE,} Sangwoo Park,~\IEEEmembership{Member,~IEEE,} Osvaldo Simeone,~\IEEEmembership{Fellow,~IEEE,} Marios Kountouris,~\IEEEmembership{Fellow,~IEEE}, David Gesbert,~\IEEEmembership{Fellow,~IEEE} % <-this % stops a space
}

% The paper headers
\markboth{}%
{Shell \MakeLowercase{\textit{et al.}}: A Sample Article Using IEEEtran.cls for IEEE Journals}

\IEEEpubid{}
% Remember, if you use this you must call \IEEEpubidadjcol in the second
% column for its text to clear the IEEEpubid mark.

\maketitle

\begin{abstract}
  Standard Bayesian learning is known to have suboptimal generalization capabilities under misspecification and in the presence of outliers. PAC-Bayes theory  demonstrates that the free energy criterion minimized by Bayesian learning is a bound on the generalization error for Gibbs predictors (i.e., for single models drawn at random from the posterior) under the assumption of sampling distributions uncontaminated by outliers. This viewpoint provides a justification for the limitations of Bayesian learning when the model is misspecified, requiring ensembling, and when data is affected by outliers. In recent work, PAC-Bayes bounds -- referred to as PAC$^m$ -- were derived to introduce free energy metrics that account for the performance of ensemble predictors, obtaining enhanced performance under misspecification. This work presents a novel robust free energy criterion that combines the generalized logarithm score function with PAC$^m$ ensemble bounds. The proposed free energy training criterion produces predictive distributions that are able to concurrently counteract the detrimental effects of misspecification {\color{black} -- with respect to both likelihood and prior distribution -- }and outliers.
\end{abstract}

\section{Introduction}
Key assumptions underlying Bayesian inference and learning are that the adopted probabilistic model is well specified  and that the training data set does not include outliers, so that training and testing distributions are matched \cite{theodoridis2015machine}.  Under these favorable conditions, the Bayesian posterior distribution provides an optimal solution to the inference and learning problems. In contrast, optimality does not extend to scenarios characterized by misspecification \cite{walker2013bayesian,grunwald2017inconsistency} \textit{or} outliers \cite{martinez2018practical}. This work aims at addressing \textit{both} problems  by integrating the use of ensemble predictors \cite{madigan1996bayesian},  generalized logarithm score functions \cite{sypherd2019tunable}, {\color{black} and generalized prior-dependent information-theoretic regularizers \cite{knoblauch2022optimization}} in Bayesian learning.

\blue{The proposed learning framework -- termed $(m,t)$-robust Bayesian learning -- is underpinned by a novel \textit{free energy} learning criterion parameterized by integer $m\geq1$ and scalar $t\in[0,1]$. The parameter $m$ controls robustness to misspecification by determining the size of the ensemble used for prediction. In contrast, parameter $t$ controls robustness to outliers by dictating the degree to which the loss function penalizes low predictive probabilities.
The proposed learning criterion generalizes the standard free energy criterion underlying Bayesian learning, which is obtained for $m=1$ and $t=1$  \cite{knoblauch2019generalized,simeone2022machine}; as well as the $m$-free energy criterion, obtained for $t=1$, which was recently introduced in \cite{morningstar2020pac}. {\color{black} A further generalization of $(m,t)$-robust Bayesian learning is also introduced, which aims at ensuring robustness to prior misspecification by modifying the information-theoretic regularizer present in the free energy \cite{knoblauch2022optimization}.}}
\begin{figure}
    \hspace{-1em}
    \centering
    \includegraphics[width=0.5\textwidth]{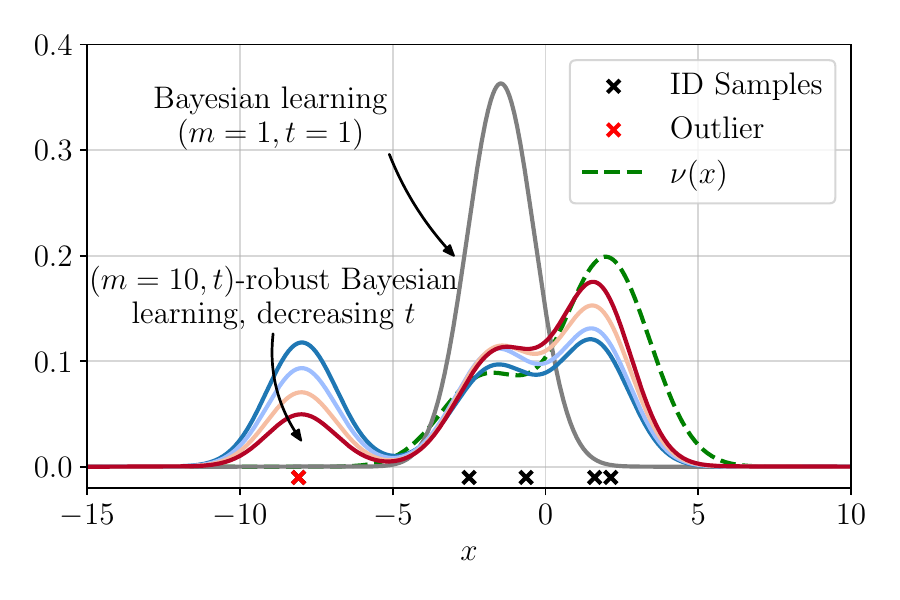}
    \caption{\blue{Ensemble predictive distributions $p_{q}(x)$ obtained via conventional and robust Bayesian learning based on the data set represented as crosses. The underlying in-distribution (ID) measure $\nu(x)$ (dashed line -- a mixture of Gaussians) produces the data points in black, while the contaminating out-of-distribution (OOD) measure $\xi(x)$ produces the data point in red. Conventional Bayesian learning  ($m=1,\ t=1$) is shown in gray; the $(m,1)$-robust Bayesian learning approach in \cite{morningstar2020pac} with $m=10$ and $t=1$ is in dark blue; and the proposed $(m,t)$-robust Bayesian learning with $m=10$ and $t=\{0.9,0.7,0.5\}$ is displayed in light blue, pink and red.}}
    \label{fig:toy_example_comparison}
\end{figure}

\blue{To illustrate the shortcomings of conventional Bayesian learning and the advantages of the proposed $(m,t)$-robust Bayesian learning, consider the example in Figure \ref{fig:toy_example_comparison}. In it, the in-distribution (ID) data generating measure (dashed line) is multimodal, while the probabilistic model is Gaussian, and hence misspecified. Furthermore, the training data set, represented as crosses, comprises an outlying data point depicted in red. In these conditions, the predictive distribution resulting from standard Bayesian learning (see the gray curve labeled as $m=1,\ t=1$) is unimodal, and it poorly approximates the underlying ID measure. 
The predictive distribution resulting from the minimization of the $m$-free energy criterion \cite{morningstar2020pac}, which corresponds to $(m,1)$-robust Bayesian learning, mitigates misspecification, being multimodal, but it is largely affected by the outlying data point (see the dark blue curve associated to $m=10, \ t=1$). Conversely, $(m,t)$-robust Bayesian learning for $t<1$ mitigates both misspecification and the presence of the outlier: the  predictive distribution resulting from the minimization of the proposed robust $(m,t)$-free energy criterion (see light blue, pink and red curves) is not only multimodal, but it can also better suppress the effect of outliers.}

\subsection{\blue{Related Work}}
\green{Recent work has addressed the problem of model misspecification for Bayesian learning, using tighter approximations of the ensemble risk \cite{masegosa2019learning,morningstar2020pac}, using pseudo-likelihoods \cite{cherief2020mmd} or modeling aleatoric uncertainty \cite{kendall2017uncertainties}. In particular, references \cite{masegosa2019learning,morningstar2020pac} have argued that the minimization of the standard free energy criterion -- which defines Bayesian learning \cite{knoblauch2019generalized,simeone2022machine} -- yields predictive distributions that do not take advantage of ensembling, and thus have poor generalization capabilities for misspecified models.}

To mitigate this problem, references \cite{masegosa2019learning,morningstar2020pac} introduced alternative free energy criteria that account for misspecification. The author of \cite{masegosa2019learning} leveraged a second-order Jensen's inequality to obtain a tighter bound on the cross entropy loss; while the work \cite{morningstar2020pac} proposed an $m$-free energy criterion that accounts for the performance of an ensemble predictor with $m$ constituent models. Both optimization criteria were shown to be effective in overcoming the shortcomings of Bayesian learning under misspecification, by yielding posteriors that make better use of ensembling. 
 
The free energy metrics introduced in  \cite{masegosa2019learning,morningstar2020pac} are defined by using the standard log-loss, which is known to be sensitive to outliers. This is because the log-loss grows unboundedly on data points that are unlikely under the model \cite{jewson2018principles}. Free energy criteria metrics based on the log-loss amount to Kullback–Leibler (KL) divergence measures between data and model distributions. A number of  papers have proposed to mitigate the effect of outliers by replacing the classical criteria based on the KL divergence in favor of more robust divergences, such as the $\beta$-divergences \cite{basu1998robust,ghosh2016robust} and the $\gamma$-divergence \cite{fujisawa2008robust,nakagawa2020robust}. These criteria can be interpreted as substituting the log-loss with generalized logarithmic scoring rules. To optimize such criteria, variational methods have been proposed that were shown to be robust to outliers, while not addressing model misspecification  \cite{futami2018variational}.

{\color{black} A separate line of work is focused on addressing the problem of prior misspecification. Prior misspecification refers to learning scenarios in which the true underlying distribution may not be well-represented by the chosen prior, causing Bayesian methods to produce biased or erroneous predictions. To address this issue, generalized formulations of Bayesian learning were introduced that aim to mitigate the influence of misspecified priors by generalizing the standard Bayesian learning criterion to alternative classes of prior-dependent information-theoretic regularizers \cite{li2016renyi,yue2019renyi,knoblauch2019generalized,knoblauch2019frequentist,knoblauch2022optimization}.}

\subsection{\blue{Contributions}}

\green{ This work extends standard Bayesian learning by concurrently tackling model misspecification, with respect to both likelihood and prior, and the presence of outliers. Specifically, the contributions of this paper are as follows.
\begin{itemize}
    \item  We introduce the $(m,t)$-robust Bayesian learning framework, which is underpinned by a novel free energy criterion based on ensemble-based loss measures and generalized logarithmic scoring rules. The predictive distribution resulting from the minimization of the proposed objective takes full advantage of ensembling, while at the same time reducing the effect of outliers.
    \item  We generalize the $(m,t)$-robust Bayesian learning framework to encompass Rényi divergence-based prior regularizers, which allow to mitigate the detrimental effect of misspecified priors.
    \item We theoretically justify and analyze the proposed robust $m$-free energy criterion within the PAC-Bayesian framework, and we prove its enhanced robustness through the lens of the influence function \cite{hampel1974influence}. 
    \item We present a wide array of experiments that corroborate the theoretical results, while also highlighting the enhanced generalization capabilities and calibration performance of the proposed learning criterion under model misspecification, prior misspecification and with data sets corrupted by outliers.
\end{itemize}}

\subsection{\blue{Paper Organization}}
The rest of the paper is organized as follows. In Section \ref{Sec2}, we review the generalized logarithm function, the associated entropy and divergence measures. We also formally describe the learning setup, providing the definition of model misspecification and introducing the contamination model under consideration. After reviewing the standard free energy criterion and its multi-sample version proposed in \cite{morningstar2020pac}, we provide a toy example highlighting the shortcoming of these two learning criteria when the model class is misspecified and the training data contains outliers. Then, in Section \ref{Sec3}, we introduce the { $(m,t)$-robust Bayesian learning framework} that tackles both model misspecification and the presence of outliers, and that overcomes the limitations of the standard Bayesian learning rule. We theoretically analyze the proposed learning criterion, providing PAC-Bayesian guarantees for the ensemble model with respect to the contaminated and the in-distribution measures. {\color{black} In Sec. \ref{sec:genBayes}, we introduce a generalized form of robust Bayesian learning that addresses also pior misspecification.} Finally, in Section \ref{Sec4}, we provide regression and classification experiments to quantitatively and qualitatively measure the performance of the proposed learning criterion.

\section{Preliminaries}

\label{Sec2}
\subsection{Generalized Logarithms}
\begin{figure}
    \centering
    \hspace{-1em}
    \includegraphics[width=0.5\textwidth]{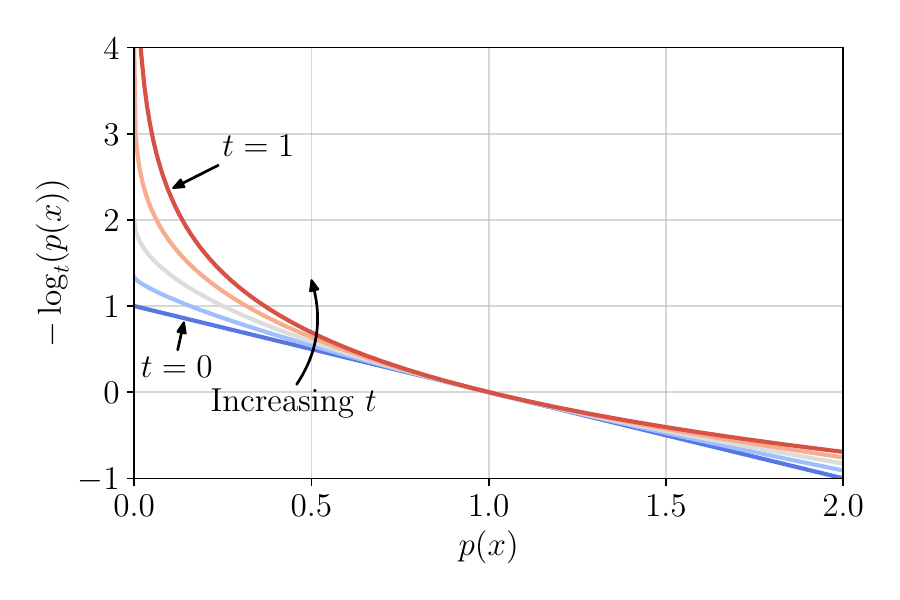}
    \caption{{$t$-logarithm loss, or $\log_t$-loss, of a predictive distribution $p(x)$ for different values of $t$. For $t=1$, the samples $x$ corresponding to low predictive probability $p(x)\to0$ have a potentially unbounded loss value. On the contrary, for $t<1$, the $t$-logarithm loss is bounded by $(1-t)^{-1}$ and it limits their influence.}}
    \label{fig:t_log_fig}
\end{figure}

The \textit{$t$}-logarithm function, also referred to as generalized or tempered logarithm is defined as
\begin{equation}
    \log_t(x):=\frac{1}{1-t}\left(x^{1-t}-1\right) \, \text{ for } x>0,
    \label{t_log_def}
\end{equation}
for $t\in[0,1)\cup(1,\infty)$, and 
\begin{equation}
   \log_1(x):=\log(x) \, \text{ for } x>0
    \label{log_def}
\end{equation}
where the standard logarithm (\ref{log_def}) is recovered from (\ref{t_log_def}) in the limit $\lim_{t\to 1}\log_t(x)=\log(x)$. As shown in Figure \ref{fig:t_log_fig}, for $t\in[0,1)$, the $t$-logarithms is a concave function, and for $t<1$ is lower bounded as $\log_t(x)\geq -(1-t)^{-1}$. 

Largely employed in classical and quantum physics, the $t$-logarithm has also been applied to machine learning problems. Specifically, $t$-logarithms have been used to define alternatives to the $\log$-loss as a score function for probabilistic predictors with the aim of enhancing robustness to outliers \cite{sypherd2019tunable,amid2018more,amid2019robust}. Accordingly, the loss associated to a probabilistic model $q(x)$ is measured as $-\log_t q(x)$ instead of the standard $\log$-loss $-\log q(x)$. Note that we have the upper bound $-\log_t q(x)\leq (1-t)^{-1}$ for $t<1$.

In information theory, the $t$-logarithm was used by  \cite{tsallis1988possible} to define the $t$-Tsallis
 entropy
\begin{align}
H_t(p(x)):=-\int p(x)^t\log_t p(x) dx, \label{TS_Ent}
\end{align}
and the $t$-Tsallis divergence
\begin{align}
D_t(p(x)||q(x)):=-\int p(x)^t[\log_t p(x)-\log_t q(x)]dx. \label{TS_Div}
\end{align}

When using the Tsallis divergence (\ref{TS_Div}) as an optimization criterion in machine learning, the concept of escort distribution is often useful \cite{sears2008generalized}. Given a probability density $p(x)$, the associated $t$-escort distribution is defined as 
\begin{align}
    \mathcal{E}_t(p(x))= \frac{p(x)^t}{\int p(x)^tdx}.
    \label{escort}
\end{align}

% \green{
% Another popular divergence related to the Tsallis divergence, and hence also to the $t$-logarithm, is the $\alpha$-Rényi entropy. For $\alpha>0$ and $\alpha\neq1$ is defined as
% \begin{align}
% H^R_{\alpha}(p(x)):=-\frac{1}{1-\alpha}\log\int p(x)^\alpha dx, \label{Renyi_Ent}
% \end{align}
% and the associated $\alpha$-Rényi divergence as 
% \begin{align}
% D^R_{\alpha}(p(x)||q(x)):=-\frac{1}{\alpha-1}\log\int p(x)^{1-\alpha}q(x)^\alpha dx. \label{Renyi_Div}
% \end{align}
% For $\alpha\to 1$  the Shannon (differential) entropy and with the Kullback–Leibler (KL) divergence are recovered.
% }

{\color{black}
Another popular divergence related to the Tsallis divergence, and hence also to the $t$-logarithm, is the $t$-Rényi entropy. For $t\in[0,1)\cup(1,\infty)$, it is defined as
\begin{align}
H^R_{t}(p(x)):=\frac{1}{1-t}\log\int p(x)^t dx, \label{Renyi_Ent}
\end{align}which can be obtained as a monotonically increasing function of the $t$-Tsallis entropy as\begin{align}
H^R_{t}(p(x))=\frac{1}{1-t}\log\left(1+(1-t) H_t(p(x))\right) . \label{Renyi_Ent}
\end{align}
The associated $t$-Rényi divergence is defined as  
\begin{align}
D^R_{t}(p(x)||q(x)):=\frac{1}{t-1}\log\int p(x)^{t}q(x)^{1-t} dx, \label{Renyi_Div}
\end{align} which is related via a monotonically increasing function to the $t$-Tsallis divergence as\begin{align}
D^R_{t}(p(x)||q(x))=\frac{1}{t-1}\log\left(1+(t-1) D_t(p(x)||q(x))\right). \label{Renyi_Ent}
\end{align}

In the limit  $t\to 1$, both $t$-Tsallis and $t$-Rényi entropies recover the Shannon (differential) entropy, i.e.,
\begin{align}
    \lim_{t\to1}H_t(p(x))=\lim_{t\to1}H^R_{t}(p(x))=\mathbb{E}_{p(x)}[-\log p(x)].
\end{align}
Furthermore, under the same limit, both $t$-Tsallis and $t$-Rényi divergences recover the Kullback–Leibler (KL) divergence, i.e., 
\begin{align}
    \lim_{t\to1}D_t(p(x)||q(x))\hspace{-0.2em}=\hspace{-0.2em}\lim_{t\to1}D^R_{t}(p(x)||q(x))\hspace{-0.2em}=\hspace{-0.2em}\mathbb{E}_{p(x)}\left[\log \frac{p(x)}{q(x)}\right]\hspace{-0.25em}.
\end{align}}

We finally note that $t$-logarithm does not satisfy the distributive property of the logarithm, i.e., $\log(xy)=\log(x)+\log(y)$. Instead,  we have the equalities \cite{umarov2008q}
\begin{align}
\log_t(xy)&=\log_t x+\log_t y +(1-t)\log_t x \log_t y
\end{align}
and
\begin{align}
\log_t \left(\frac{x}{y}\right)&=y^{t-1}\left(\log_t x-\log_t y\right).
\end{align}

\subsection{Assumptions and Motivation}
\label{sec:setup}

\begin{table}[]
\centering
\caption{Total variation (TV) distance between the ID measure $\nu(x)$ and the predictive distribution $p_q(x)$ obtained from the optimization of the different free energy criteria described in the text.}
\begin{tabular}{@{}llllll@{}}
\toprule
   & \begin{tabular}{@{}l@{}}$m=1$ \\ $t=1$\end{tabular}  & \begin{tabular}{
   @{}l@{}}$m=10$ \\ $t=1$\end{tabular}   &  \begin{tabular}{@{}l@{}}$m=10$ \\ $t=0.9$\end{tabular}  & \begin{tabular}{@{}l@{}}$m=10$ \\ $t=0.7$\end{tabular}  & \begin{tabular}{@{}l@{}}$m=10$ \\ $t=0.5$\end{tabular} \\ \midrule
$\textrm{TV}(\nu(x)||p_q(x))$ \hspace{-1em} & 0.59  & 0.35                   & 0.30 & 0.24 & 0.19
\end{tabular}
\label{tab:tv_dist_fig1}
\end{table}
We consider a learning setup in which the data distribution is contaminated by outliers \cite{Huber}, and the assumed parametric model is misspecified \cite{masegosa2019learning}. As in \cite{Huber}, the presence of outliers is modelled by assuming that the observed vector $x\in \mathcal{X}$ follows a sampling distribution $\tilde{\nu}(x)$ given by the contamination of an in-distribution (ID) measure $\nu(x)$ by an out-of-distribution (OOD) measure $\xi(x)$. 
\begin{assumption}[Outliers]
\label{ass:out}
The sampling distribution follows the gross error model proposed by \cite{Huber},
\begin{align}
    \tilde{\nu}(x)=(1-\epsilon)\nu(x)+\epsilon\xi(x)
    \label{cont_mod}
\end{align}
where $\nu(x)$ is the ID measure; $\xi(x)$ is the OOD measure accounting for outliers; and $\epsilon\in[0,1]$ denotes the contamination ratio.
\end{assumption} 

In order for model (\ref{cont_mod}) to be meaningful, one typically assumes that the OOD measure $\xi(x)$ is large for values of $x$ at which the ID measure $\nu(x)$ is small. 

The learner is assumed to have access to a data set $\mathcal{D}=\{(x_i)\}^n_{i=1}\sim \tilde{\nu}(x)^{\otimes n}$ drawn from the sampling distribution, and it assumes a uniformly upper bounded parametric model family $p_{\theta}(x)$ defined by a model parameter vector $\theta\in \Theta$, which is generally misspecified. 
\begin{assumption}[Misspecification]
\label{ass:mis}
The model class $\{p_{\theta}(\cdot):\theta\in\Theta\}$ is misspecified with respect to the ID measure $\nu(x)$, in the sense that there is no model parameter vector $\theta\in \Theta$ such that $\nu(x)=p_\theta(x)$. Furthermore, it is uniformly upper bounded, in the sense that there exists a finite constant $C$ such that $p_\theta(x)\leq C$ for all $\theta\in\Theta$ and values of $x\in\mathcal{X}$.
\end{assumption}
In order to account for misspecification, as in \cite{masegosa2019learning}, we adopt ensemble models of the form 
\begin{align}
    p_q(x):=\mathbb{E}_{q(\theta)}[p_\theta(x)], \label{ens_dist}
\end{align}
where $q(\theta)$ is the ensembling distribution on the model parameter space $\Theta$. The rationale for considering ensemble models is that the average (\ref{ens_dist}), which accounts from the combinations of multiple models $p_\theta(\cdot)$, may better represent the ID measure $\nu(x)$ in the presence of misspecification (see Assumption \ref{ass:mis}). 

The $\log_t$-loss of the ensemble model (\ref{ens_dist}) is given as 
\begin{align}
     \mathcal{R}_t(q,x):=-\log_t p_q(x)=-\log_t \mathbb{E}_{q(\theta)}[p_\theta(x)].
     \label{ens_risk_x}
\end{align}
This contrasts with the average $\log_t$-loss obtained by drawing a model parameter vector $\theta\sim q(\theta)$ and then using the resulting model $p_{\theta}(x)$ --- an approach known as Gibbs model. The corresponding $\log_t$-loss is
\begin{align}
     \hat{\mathcal{R}}_t(q,x):=\mathbb{E}_{q(\theta)}[-\log_t p_\theta(x)].
     \label{Gibbs_risk_x}
\end{align}
Unlike the standard $\log$-loss with $t=1$, the $\log_t$-loss is upper bounded by $(1-t)^{-1}$ for any $t\in(0,1)$. This constrains the impact of anomalous data points to which the model --- either $p_{\theta}(x)$ or $p_q(x)$ for Gibbs and ensemble predictors, respectively --- assigns low probability.

Since the sampling distribution $\tilde{\nu}(x)$ is not known, the risk 
\begin{align}
     \mathcal{R}_t(q):=\mathbb{E}_{\tilde{\nu}(x)}[-\log_t \mathbb{E}_{q(\theta)}[p_\theta(x)]] \label{ens_risk}
\end{align}
of the ensemble model cannot be computed by the learner. However, for $t=1$, using Jensen's inequality and standard PAC-Bayes arguments, the risk (\ref{ens_risk}) can be upper bounded using the data set $\mathcal{D}$ (and neglecting inessential constants) by the free energy criterion \cite{theodoridis2015machine,catoni2003pac}
\begin{align}
    \mathcal{J}(q):= \frac{1}{n}\sum_{x\in\mathcal{D}}\hat{\mathcal{R}}_1(q,x)+\frac{D_{1}(q(\theta)||p(\theta))}{\beta} \label{van_PAC}
\end{align}
where we recall that $D_{1}(q(\theta)||p(\theta))$ is the KL divergence with respect to a prior distribution $p(\theta)$, while $\beta>0$ is a constant, also known as inverse temperature.

The criterion ($\ref{van_PAC}$), for $\beta=n$, is minimized by the standard Bayesian posterior i.e.,
\begin{align}
q_{Bayes}(\theta)&=\argmin_q \sum_{x\in\mathcal{D}}\hat{\mathcal{R}}_1(q,x)+D_{1}(q(\theta)||p(\theta))\\
&\propto p(\mathcal{D}|\theta)p(\theta).
\end{align}
Even disregarding outliers, in the misspecified setting, the resulting ensemble predictor $p_{q_{Bayes}}(x)=\mathbb{E}_{q_{Bayes}(\theta)}[p_\theta(x)]$ is known to lead to poor performance, as the criterion (\ref{van_PAC}) neglects the role of ensembling to mitigate misspecification \cite{masegosa2019learning}. 

\textit{Example:} Consider a Gaussian model class $p_{\theta}(x)=\mathcal{N}(x|\theta,1)$ and a prior $p(\theta)=\mathcal{N}(\theta|0,9)$. We obtain the standard Bayesian posterior $q_{Bayes}(\theta)$ by minimizing the free energy (\ref{van_PAC}) with $n=5$  data points sampled from the ID measure $\nu(x)= 0.7\mathcal{N}(x|2,2)+0.3\mathcal{N}(x|-2,2)$, contaminated by an OOD measure $\xi(x)=\mathcal{N}(x|-8,1)$ with a contamination ratio $\epsilon=0.1$. Note that the model is misspecified, since it assumes a single Gaussian, while the ID measure $\nu(x)$ is a mixture of Gaussians. Therefore, the resulting predictive ensemble distribution $p_{q_{Bayes}}(x)$ (gray line) is not able to capture the multimodal nature of the sampling distribution. Furthermore, the presence of outliers leads to a predictive distribution that deviates from the ID distribution $\nu(x)$ (green dashed line).\hfill $\blacksquare$

\subsection{PAC$^m$-Bayes}
The limitations of the standard PAC-Bayes risk bound (\ref{van_PAC}) as a learning criterion in the presence of model misspecification have been formally investigated by \cite{masegosa2019learning} and \cite{morningstar2020pac}. These works do not consider the presence of outliers, hence setting the sampling distribution $\tilde{\nu}(x)$ to be equal to the ID measure $\nu(x)$ (or $\epsilon=0$ in \ref{cont_mod}). Here we review the PAC$^m$ bound introduced by  \cite{morningstar2020pac} with the goal of overcoming the outlined limitations of (\ref{van_PAC}) in the presence of misspecification.

In \cite{morningstar2020pac}, the free energy criterion ($\ref{van_PAC}$) is modified by replacing the Gibbs risk $\hat{\mathcal{R}}_1(q,x)$  with a sharper bound on the ensemble $\log$-loss $\mathcal{R}_1(q,x)$. For $m\geq1$, this bound is defined as
 \begin{align}
     \hat{\mathcal{R}}_1^m(q,x)\hspace{-0.2em}:=\hspace{-0.2em}\mathbb{E}_{\theta_1,\dots,\theta_m\sim q(\theta)^{\otimes m}}\left[-\log\mathbb{E}_{j\sim U[1:m]} [p(x|\theta_j)]\right]\hspace{-0.2em} \label{mutlisample_log}
\end{align}
where the inner expectation is over an index $j$ uniformly distributed in the set $[1:m]=\{1,2,\dots,m\}$.

By leveraging the results of \cite{burda2016importance} and \cite{mnih2016variational}, the multi-sample criterion $\hat{\mathcal{R}}_1^m(q,x)$ can be shown to provide a sharper bound to the ensemble risk $\mathcal{R}_1(q,x)$ in (\ref{ens_risk_x}) as compared to the Gibbs risk  $\hat{\mathcal{R}}_1^1(q,x)$ in (\ref{Gibbs_risk_x}), i.e.,
 \begin{align}
     \mathcal{R}_1(q,x) \leq \hat{\mathcal{R}}_1^m(q,x)\leq  \hat{\mathcal{R}}_1^1(q,x)=\hat{\mathcal{R}}_1(q,x) \label{chain_ineq}
\end{align}
Furthermore, the first inequality in (\ref{chain_ineq}) becomes asymptotically tight as $m\to \infty$, i.e.,
\begin{align} 
    \lim_{m\to\infty} \hat{\mathcal{R}}_1^m(q,x)=\mathcal{R}_1(q,x). \label{lim_inf}
\end{align}
Using PAC-Bayes arguments, \cite{morningstar2020pac} show that the $\log$-risk (\ref{ens_risk}) with $t=1$ and $\tilde{\nu}(x)=\nu(x)$ can be upper bounded, for $\beta>0$, (neglecting inessential constants) by the $m$-free energy criterion 
\begin{align}
    \mathcal{J}^m(q):=\frac{1}{n} \sum_{x\in\mathcal{D}}\hat{\mathcal{R}}_1^m(q,x)+\frac{m}{\beta} D_1(q(\theta))||p(\theta)).
    \label{eq:morningstar_proposed_obj}
\end{align}

The minimization of the $m$-free energy $\mathcal{J}^m(q)$ produces a posterior
\begin{align}
    q^m(\theta):=\argmin_q \mathcal{J}^m(q),
\end{align}
which can take better advantage of ensembling, resulting in predictive distributions that are more expressive than the ones obtained following the standard Bayesian approach based on (\ref{van_PAC}). 

\textit{Example (continued):} This is shown in Figure \ref{fig:toy_example_comparison}, in which we plot the predictive distribution $p_{q^m}(\theta)$ obtained by minimizing the $m$-free energy $\mathcal{J}^m(q)$ in (\ref{eq:morningstar_proposed_obj}) for $m=10$ for the same example described in the previous subsection. The optimized predictive distribution $p_{q^m}(x)$ is multimodal; it covers all data samples; and, as shown in Table \ref{tab:tv_dist_fig1}, it reduces the total variational distance from the ID measure $\nu(x)$ as compared to the predictive distribution obtained minimizing $\mathcal{J}(q)$. \hfill $\blacksquare$

\section{$(m,t)$-Robust Bayesian Learning}
\label{Sec3}
In the previous section, we reviewed the $m$-free energy criterion introduced by \cite{morningstar2020pac}, which was argued to produce predictive distributions that are more expressive, providing a closer match to the underlying sampling distribution $\nu(x)$. However, the approach is not robust to the presence of outliers. In this section, we introduce { $(m,t)$-robust Bayesian learning and the associated novel free energy criterion} that addresses both expressivity in the presence of misspecification and robustness in setting with outliers. To this end, we study the general setting described in Section \ref{sec:setup} in which the sampling distribution $\tilde{\nu}(x)$ satisfies both Assumption \ref{ass:out} and Assumption \ref{ass:mis}, and we investigate the use of the $\log_t$-loss with $t\in[0,1)$ as opposed to the standard $\log$-loss as assumed in \cite{morningstar2020pac}.

\subsection{Robust $m$-free Energy}
For a proposal posterior $q(\theta)$, generalizing (\ref{mutlisample_log}), we define the multi-sample empirical $\log_t$-loss evaluated at a data point $x$ as
 \begin{align}
     \hat{\mathcal{R}}^m_t(q,x)\hspace{-0.2em}:=\hspace{-0.2em}\mathbb{E}_{\theta_1,\dots,\theta_m\sim q(\theta)^{\otimes m}}\left[-\log_t\hspace{-0.2em}\mathbb{E}_{j\sim U[1:m]}p(x|\theta_j)\right]. \label{emp_t_obj}
\end{align}
From the concavity of the $t$-logarithm with $t\in[0,1)$, in a manner similar to (\ref{chain_ineq}), the loss (\ref{emp_t_obj}) provides an upper bound on the original $\log_t$-loss  $\mathcal{R}_t(q,x)$ in (\ref{ens_risk_x})
\begin{align}
 \mathcal{R}_t(q,x)\leq \hat{\mathcal{R}}^m_t(q,x).
\end{align}
Furthermore, the bound becomes increasingly tighter as $m$ increases, and we have the limit
\begin{align}
    \lim_{m\to\infty}\hat{\mathcal{R}}^m_t(q,x)= \mathcal{R}_t(q,x)
\end{align}
for $t\in[0,1)$. The $m$-sample $\log_t$-loss (\ref{emp_t_obj}) is used to define, for $\beta>0$, the robust $m$-free energy as  
\begin{align}
    \mathcal{J}^m_t(q):=\frac{1}{n}\sum_{x\in\mathcal{D}}\hat{\mathcal{R}}^m_t(q,x)+\frac{m}{\beta} D_1(q(\theta)||p(\theta)).
    \label{eq:proposed_obj}
\end{align}
{The proposed free energy generalizes the standard free energy criterion (\ref{van_PAC}), which corresponds to the training criterion of $(m,t)$-robust Bayesian learning for $m=1$ and $t=1$, and the $m$-free energy criterion (\ref{eq:morningstar_proposed_obj}), which corresponds to the training criterion  of  $(m,t)$-robust Bayesian learning for $t=1$.}

Following similar steps as in \cite{morningstar2020pac}, the robust $m$-free energy can be proved to provide an upper bound on the $\log_t$-risk in (\ref{ens_risk}), as detailed in the following lemma.
\begin{lemma}
\label{lem:rob_pac}
With probability $1-\sigma$, with $\sigma\in(0,1)$, with respect to the random sampling of the data set $\mathcal{D}$, for all distributions $q(\theta)$ that are absolutely continuous with respect the prior $p(\theta)$, the following bound on the risk ($\ref{ens_risk}$) of the ensemble model holds
\begin{align}
    \mathcal{R}_t(q)\leq&\mathcal{J}^m_t(q)+\psi(\tilde{\nu},n,m,\beta,p,\sigma)
     \label{Rob_PAC_samp}
\end{align}
where 
\begin{align}
   \psi(\tilde{\nu},n,m,\beta,p,\sigma):=\frac{1}{\beta}\left(\log\mathbb{E}_{\mathcal{D},p(\theta)}\left[e^{\beta\Delta_{m,n}}\right]-\log\sigma\right)
\end{align}
and
\begin{align}
   \Delta_{m,n}:=&\frac{1}{n}\sum_{x\in \mathcal{D}}\log_t\hspace{-0.2em}\mathbb{E}_{j\sim U[1:m]}p(x|\theta_j)\nonumber\\
   &-\mathbb{E}_{\tilde{\nu}(x)}\left[\log_t\hspace{-0.2em}\mathbb{E}_{j\sim U[1:m]}p(x|\theta_j)\right].
\end{align}
Furthermore, the risk with respect to the ID measure $\nu(x)$ can be bounded as 
\begin{align}
    \mathbb{E}_{\nu(x)}[\mathcal{R}_t(q,x)]\leq&\frac{1}{1-\epsilon}\left( \mathcal{J}^m_t(q)+\psi(\tilde{\nu},n,m,\beta,p,\sigma)\right)\nonumber\\
    &+\frac{\epsilon(C^{1-t}-1)}{(1-\epsilon)(1-t)},
    \label{Rob_PAC_clean}
\end{align}
if the contamination ratio satisfies the inequality $\epsilon<1$.
\end{lemma}

Lemma \ref{lem:rob_pac} provides an upper bound on the $\log_t$-risk (\ref{ens_risk}), which is defined with respect to the sampling distribution $\tilde{\nu}(x)$ corrupted by outliers, as well as on the ensemble $\log_t$-risk (\ref{ens_risk_x}) evaluated with respect to the ID measure $\nu(x)$. Reflecting that the data set $\mathcal{D}$ contains samples from the corrupted measure $\tilde{\nu}(x)$, while the bound (\ref{Rob_PAC_samp}) vanishes as $n\to\infty$, a non-vanishing term appears in the bound (\ref{Rob_PAC_clean}).

\subsection{Minimizing the Robust $m$-free Energy}

Using standard tools from calculus of variations, it is possible to express the minimizer of the robust $m$-free energy 
\begin{align}
q_t^m(\theta):=\argmin_q \mathcal{J}^m_t(q)
\label{eq:minimizer_proposed}
\end{align}
as fixed-point solution of an operator acting on the ensembling distribution $q(\theta)$.
\begin{thm}
\label{thm1}
The minimizer (\ref{eq:minimizer_proposed}) of the robust $m$-free energy objective (\ref{eq:proposed_obj}) is the fixed point of the operator
\begin{align}
T(q)\hspace{-0.2em}:=\hspace{-0.2em}p(\theta_j)\exp\left(\hspace{-0.2em}\beta\hspace{-0.2em}\sum_{x\in\mathcal{D}}\mathbb{E}_{\{\theta_i\}_{i\neq j}}\hspace{-0.3em}\left[\log_t\hspace{-0.2em}\left(\frac{\sum^m_{i=1} p_{\theta_i}(x)}{m}\right)\right]\hspace{-0.2em}\right)\label{minimizer_rob}
\end{align}
where the average in (\ref{minimizer_rob}) is taken with respect to the i.i.d. random vectors $ \{\theta_i\}_{i\neq j}\sim q(\theta)^{\otimes m-1}$.
\end{thm}
Theorem \ref{thm1} is useful to develop numerical solutions to problem (\ref{eq:minimizer_proposed}) for non-parametric posteriors, and it resembles standard mean-field variational inference iterations \cite{bishop2006pattern}. 

Alternatively, we can tackle the problem (\ref{eq:minimizer_proposed}) over a parametric family of distribution using standard tools from variational inference \cite{blei2017variational}.

To further characterize the posterior minimizing the robust $m$-free energy criterion, and to showcase the beneficial effect of the generalized logarithm, we now consider the asymptotic regime in which $m\to \infty$ and then $n\to\infty$. In this limit, the robust $m$-free energy (\ref{eq:proposed_obj}) coincides with the $\log_t$-risk  $\mathcal{R}_t(q)$. From the definition of $t$-Tsallis divergence (\ref{TS_Div}), the $\log_t$-risk can be shown in turn to be equivalent to the minimization of the divergence
\begin{align}
   D_t\left(\mathcal{E}_t(\tilde{\nu}(x))|| p_{q({\theta})}(x)\right) \label{Ts_div_min}
\end{align}
between the $t$-escort distribution (\ref{escort}) associated to the sampling distribution $\tilde{\nu}(x)$ and the ensemble predictive distribution $p_{q({\theta})}(x)$. Therefore, unlike the standard Bayesian setup with $t=1$, the minimizer of the robust $m$-free energy does not seek to approximate the sampling distribution $\tilde{\nu}(x)$. Instead, the minimizing ensembling posterior $q(\theta)$ aims at matching the $t$-escort version of the sampling distribution $\tilde{\nu}(x)$. In the case of corrupted data generation procedures, i.e., when $\nu(x)\neq\tilde{\nu}(x)$, recovering the sampling distribution $\tilde{\nu}(x)$ is not always the end goal, and, as shown by  \cite{sypherd2019tunable}, escort distributions are particularly effective at reducing the contribution of OOD measures. 
 
\textit{Example (continued):} Consider again the example in Figure \ref{fig:toy_example_comparison}. The minimization of the proposed robust $m$-free energy $\mathcal{J}^m_t(q)$ for $m=10$ and $t=\{0.9,0.7,0.5\}$ is seen to lead to expressive predictive distributions (\ref{eq:minimizer_proposed}) that are also able to downweight the contribution of the outlying data point. This is quantified by reduced total variation distances as seen in Table \ref{tab:tv_dist_fig1}.

\subsection{Influence Function Analysis}
\label{sec:if}
\begin{figure}
    \centering
    \hspace{-1em}
    \includegraphics[width=0.5\textwidth]{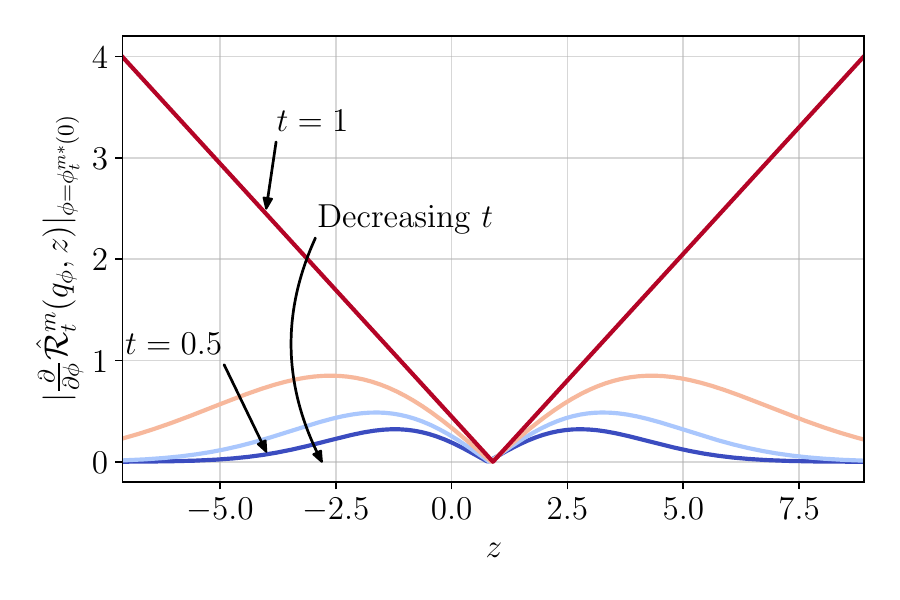}
    \caption{Absolute value of the contamination dependent term $\pdv{}{\phi}\hat{\mathcal{R}}^m_t(q_{\phi},z)$ evaluated at $\phi^{m*}_t(0)$ for different values of $t$. The predictive distribution of the ensemble model concentrates around $1$.}
    \label{fig:IFas}
\end{figure}
In this section, we study the robustness of the proposed free energy criterion by using tools from classical statistics. The robustness of an estimator is typically measured by the means of its influence function \cite{hampel1974influence}. The influence function quantifies the extent to which an estimator derived from a data set $\mathcal{D}$ changes when a data point $z$ is added to $\mathcal{D}$.
We are specifically interested in quantifying the effect of data contamination, via the addition of a point $z$, on the ensembling distribution $q_t^m(\theta)$ that minimizes the proposed robust $m$-free energy objective (\ref{eq:proposed_obj}). To this end, given a set $\mathcal{D}$ of $n$ data points $\{x_1,\dots,x_n\}\in\mathcal{X}^n$, we define the empirical measure 
\begin{align}
P^n(x)=\frac{1}{n}\sum^n_{i=1}\delta(x-x_i)
\end{align}
where $\delta(\cdot)$ denotes the Dirac function, and we introduce its $\gamma$-contaminated version for an additional data point $z\in\mathcal{X}$ as
\begin{align}
P^n_{\gamma,z}(x)=\frac{(1-\gamma)}{n}\sum^n_{i=1}\delta(x-x_i)+\gamma \delta(x-z)
\label{contaminated}
\end{align}
with $\gamma\in[0,1]$. 

The following analysis is inspired by \cite{futami2018variational}, which considered Gibbs models trained using generalized free energy criteria based on the $\beta$-divergence and $\gamma$-divergence.

To compute the influence function we consider parametric ensembling distributions $q_{\phi}(\theta)$ defined by the parameter vector $\phi\in\Phi\subseteq\mathbb{R}^d$. We denote the robust $m$-free energy (\ref{eq:proposed_obj}) evaluated using the empirical distribution (\ref{contaminated}) as
\begin{align}
    \mathcal{J}^m_t(\gamma,\phi)\hspace{-0.2em}=&\mathbb{E}_{ P^n_{\gamma,z}(x)}\hspace{-0.2em}\left[\hat{\mathcal{R}}^m_t(q_{\phi},x)\right]\hspace{-0.2em}+\hspace{-0.2em}\frac{m}{\beta} D_1(q_{\phi}(\theta)||p(\theta)), \label{PAC-t contaminated}
\end{align}
and its minimizer as
\begin{align}
\phi^{m*}_t(\gamma)=\argmin_{\phi\in\Phi} \mathcal{J}^m_t(\gamma,\phi).
\end{align}
The influence function is then defined as the derivative
\begin{align}
    IF^m_t(z,\phi,P^n)&=\frac{d\phi^{m*}_t(\gamma)}{d\gamma}\Bigg|_{\gamma=0}\label{IF_derivative}\\&=\lim_{\gamma\to 0} \frac{\phi^{m*}_t(\gamma)-\phi^{m*}_t(0)}{\gamma}.
\end{align}
Accordingly, the influence function measures the extent to which the minimizer $\phi^{m*}_t(\gamma)$ changes for an infinitesimal perturbation of the data set.
\begin{thm}
\label{IF_th}
The influence function of the robust $m$-free energy objective (\ref{PAC-t contaminated}) is
\begin{align}
IF^m_t\hspace{-0.1em}(\hspace{-0.1em}z,\phi,P^n\hspace{-0.1em})&\hspace{-0.2em}=\hspace{-0.2em}-\hspace{-0.2em}\left[\hspace{-0.1em}\pdv[2]{\mathcal{J}^m_t(\gamma,\phi)}{\phi}\hspace{-0.1em}\right]^{\hspace{-0.1em}-1}\hspace{-1.2em}\times\hspace{-0.2em}\pdv[2]{\mathcal{J}^m_t(\gamma,\phi)}{\gamma}{\phi}\hspace{-0.1em}\Bigg|_{\substack{\gamma=0\hfill\\\phi=\phi^{m*}_t\hspace{-0.2em}(0)}}\hspace{-0.4em},
\end{align}
where
\begin{align}
\pdv[2]{\mathcal{J}^m_t(\gamma,\phi)}{\phi}\hspace{-0.2em}=&\mathbb{E}_{P^n_{\gamma,z}(x)}\pdv[2]{}{\phi}\left[\hat{\mathcal{R}}^m_t(q_{\phi},x)\right]\\&+\pdv[2]{}{\phi}\left[\frac{m}{\beta} KL(q_{\phi}(\theta)||p(\theta))\right]
\end{align}
and
\begin{align}
\pdv[2]{\mathcal{J}^m_t(\gamma,\phi)}{\gamma}{\phi}\hspace{-0.2em}=\hspace{-0.2em}\pdv{}{\phi}\left[\mathbb{E}_{ P^n(x)}\left[\hat{\mathcal{R}}^m_t(q_{\phi},x)\right]\hspace{-0.2em}-\hspace{-0.2em}\hat{\mathcal{R}}^m_t(q_{\phi},z)\right].
\end{align}
\end{thm}
Theorem \ref{IF_th} quantifies the impact of the data point $z$ through the contamination dependent term $\pdv{}{\phi}\hat{\mathcal{R}}^m_t(q_{\phi},z)$.
We study the magnitude of this term to illustrate the enhanced robustness deriving from the proposed robust $m$-free energy objective. For ease of tractability, we consider the limit $m\to \infty$. In this case, the contamination dependent term can be expressed as
\begin{align}
    \pdv{}{\phi}\lim_{m\to\infty}\hspace{-0.4em}\hat{\mathcal{R}}^m_t(q_{\phi},z)\hspace{-0.2em}&=\hspace{-0.2em}\pdv{}{\phi}\log_t\mathbb{E}_{q_\phi(\theta)}[p(z|\theta)]\\
    &=\hspace{-0.2em}\left[\mathbb{E}_{q_\phi(\theta)}[p(z|\theta)]\right]^{-t}\pdv{\mathbb{E}_{q_\phi(\theta)}[p(z|\theta)]}{\phi}.
    \label{cont_term_minf}
\end{align}
The effect of the $t$-logarithm function thus appears in the first multiplicative term, and it is the one of reducing the influence of anomalous data points to which the  ensemble predictive distribution $p_q(x)$ assigns low probability.

\textit{Example:} To illustrate how the $t$-logarithm improves the robustness to outlying data points, we consider again the example of Figure \ref{fig:toy_example_comparison} and we assume a parametrized ensembling posterior $q_{\phi}(\theta)=\mathcal{N}(\theta|\phi,1)$. In Figure \ref{fig:IFas}\Note{,} we plot the magnitude of the contamination dependent term evaluated at the parameter $\phi^{m*}_t(0)$ that minimizes the robust $m$-free energy $\mathcal{J}^m_t(0,\phi)$ for $m=\infty$ and different values of $t$. For all values of $t$, the optimized predictive distribution concentrates around $0$, where most of sampled data points lie. However, as the value of the contaminated data point $z$ becomes smaller and moves towards regions where the ensemble assign low probability, the contamination dependent term grows linearly for $t=1$, while it flattens for $t\in(0,1)$. This showcases the role of the robust $m$-free energy criterion as a tool to mitigate the influence of outlying data points by setting $t<1$.
\begin{figure*}
    \centering
    \subcaptionbox{\label{1a}$\epsilon=0$, $t=1$}
        [0.245\textwidth]{ \includegraphics[width=0.24\textwidth]{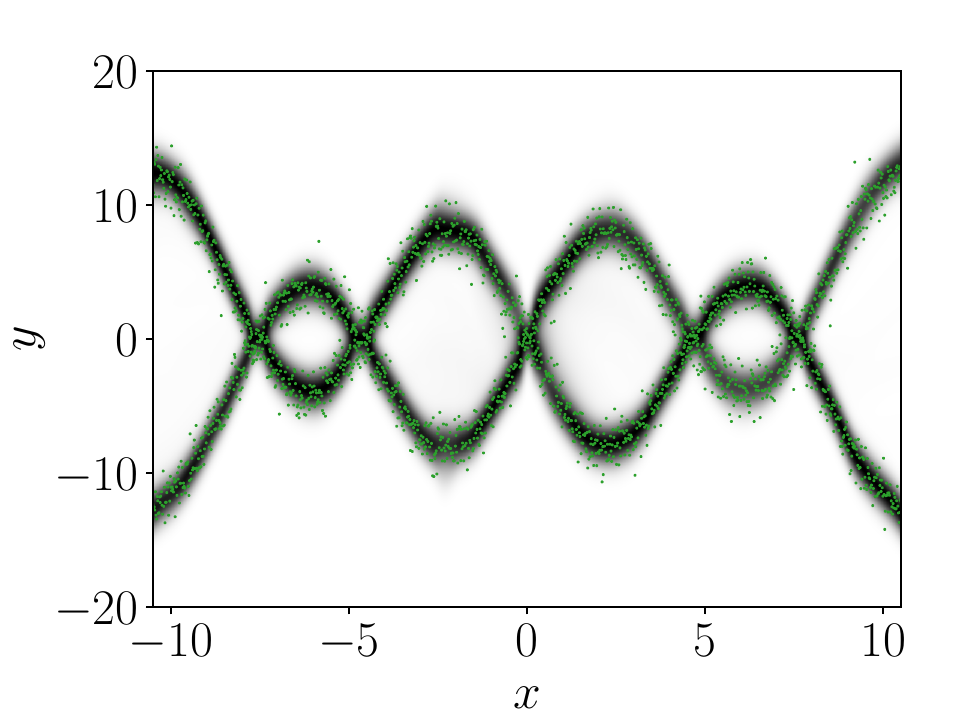}}
   \subcaptionbox{\label{1b}$\epsilon=0.1$, $t=1$}
        [0.245\textwidth]{ \includegraphics[width=0.24\textwidth]{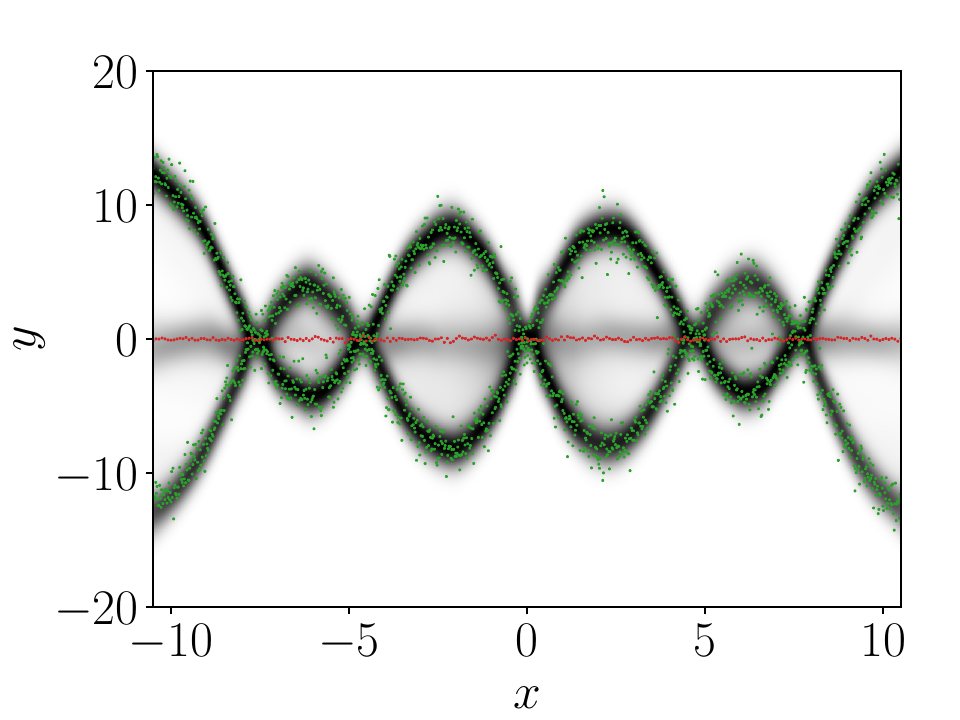}}
    \subcaptionbox{\label{1c}$\epsilon=0.1$, $t=0.9$}
        [0.245\textwidth]{ \includegraphics[width=0.24\textwidth]{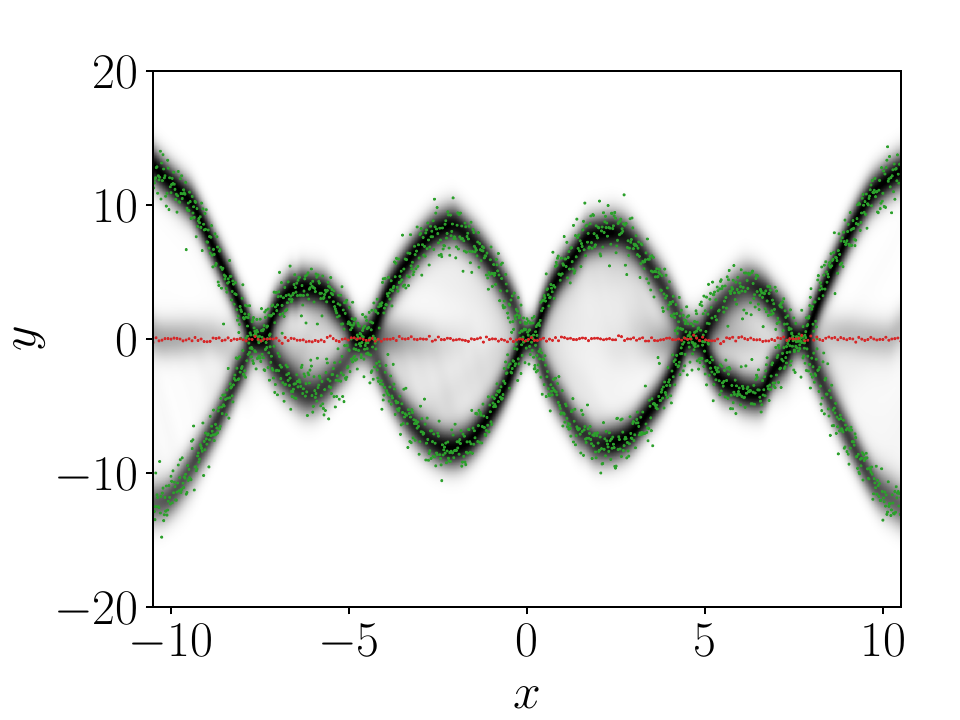}}
    \subcaptionbox{\label{1d}$\epsilon=0.1$, $t=0.8$}
        [0.245\textwidth]{ \includegraphics[width=0.24\textwidth]{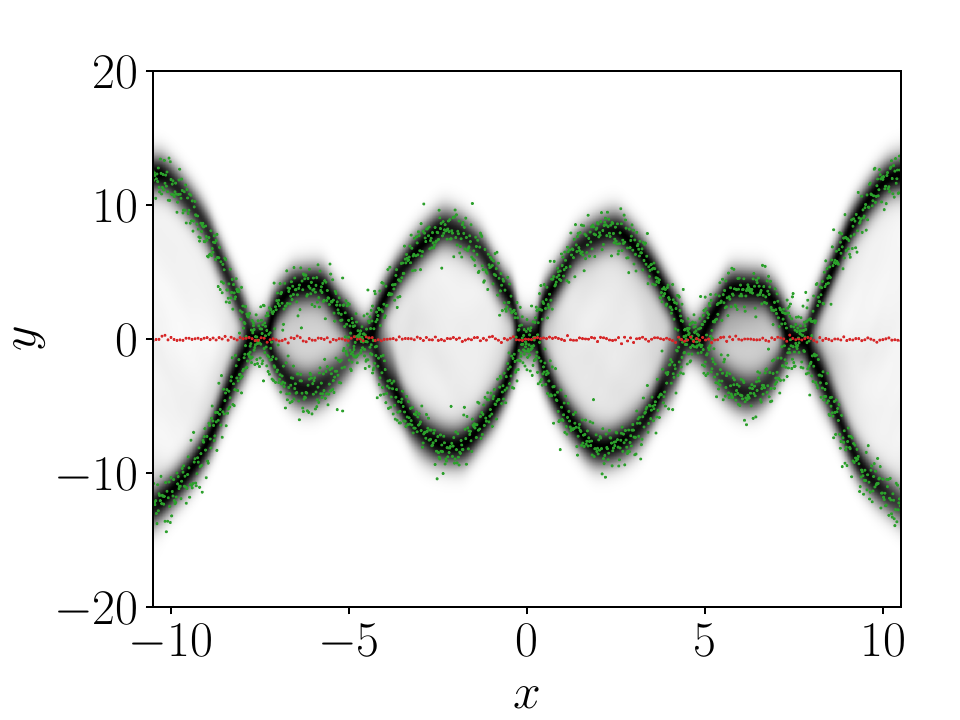}}
    \caption{Ensemble predictive distribution obtained minimizing different free energy criteria. The samples from the ID measure are represented as green dots, while data points sampled from the OOD component are in red. The optimized predictive distributions are displayed in shades of gray. In (a), we plot the predictive distribution associated to {$(m,1)$-robust Bayesian learning} obtained minimizing the $m$-free energy criterion $\mathcal{J}^m$ of \cite{morningstar2020pac} with $m=20$ by using only samples from the ID measure (i.e., there are no outliers). In (b), we show the predictive distribution obtained by minimizing the same criterion when using samples from the ID measure and OOD measure with a contamination ratio $\epsilon=0.1$. In (c) and (d) we consider the same scenario as in (b), but we consider the {proposed $(m,t)$-robust Bayesian} based on the robust $m$-free energy criterion $\mathcal{J}_t^m$ with $m=20$, when setting $t=0.9$ and $t=0.8$, respectively.}
    \label{fig:first_exp}
\end{figure*}
\begin{table}
\centering
\caption{Total variation (TV) distance between the ID measure $\nu(x)$ and the predictive distribution $p_q(x)$ obtained from the optimization of the different free energy criteria for the setting in Figure 4 (the TV values are scaled by $10^{4}$).}
\begin{tabular}{@{}llllll@{}}
\toprule
   &  \begin{tabular}{@{}l@{}} $t=1$\\ $\epsilon=0$\end{tabular} &  \begin{tabular}{@{}l@{}} $t=1$\\ $\epsilon=0.1$\end{tabular} &   \begin{tabular}{@{}l@{}} $t=0.9$\\ $\epsilon=0.1$\end{tabular} & \begin{tabular}{@{}l@{}} $t=0.8$\\ $\epsilon=0.1$\end{tabular} \\ \midrule
$\textrm{TV}(\nu(x)||p_q(x))$ & $1.38$  & $2.15$               & $1.88$ & $1.79$
\end{tabular}
\label{tab:first_exp}
\end{table}

{\color{black}
\section{Generalized $(m,t)$-Robust Bayesian Learning}\label{sec:genBayes}
    
So far, we have addressed the problem of model misspecification with respect to the likelihood function $p_\theta(\cdot)$ as defined in Assumption \ref{ass:mis}. When applying Bayesian learning, a further common concern with regards to misspecification has to do with the choice of the prior distribution $p(\theta)$. In Bayesian learning, as well as robust Bayesian learning as presented in this paper, the prior distribution  $p(\theta)$ is accounted for in the design problem  by including a regularizer $ D_1(q(\theta)||p(\theta))$ on the ensembling distribution $q(\theta)$ under optimization on the free energy objective (see (\ref{van_PAC}) for conventional Bayesian learning and (\ref{eq:proposed_obj}) for robust Bayesian learning). It has been recently argued that the KL divergence $ D_1(q(\theta)||p(\theta))$ may not offer the best choice for the regularizer when the prior is not well specified due to its mode-seeking behavior  \cite{knoblauch2022optimization,minka2005divergence,simeone2022machine}. In this section, we extend the generalized Bayesian learning framework in \cite{knoblauch2022optimization} to incorporate robustness to likelihood misspecification and outliers.

To this end, we extend the $(m,t)$-robust Bayesian learning criterion (\ref{eq:proposed_obj}) by replacing the KL divergence $ D_1(q(\theta)||p(\theta))$ with the more general $t$-Rényi divergence $ D_{t}^{R}(q(\theta)||p(\theta))$ in (\ref{Renyi_Div}). Recall that the Rényi divergence tends to the KL divergence as $t$ approaches to 1. This extension is motivated by the fact that, for $t<1$, the Rényi divergence exhibits a mass-covering behavior that has been shown to improve robustness against ill-specified prior distributions \cite{knoblauch2022optimization,li2016renyi,yue2019renyi}.

Accordingly, the generalized  $(m,t)$-robust Bayesian learning criterion is defined as  
\begin{align}
    \mathcal{J}^m_{t,t_p}(q):=\frac{1}{n}\sum_{x\in\mathcal{D}}\hat{\mathcal{R}}^m_t(q,x)+\frac{m}{\beta} D^R_{t_p}(q(\theta)||p(\theta)).
    \label{eq:gen_proposed_obj}
\end{align}
We emphasize that the parameter $t_p$ specifying the Rényi regularizer need not equal the parameter  $t$ used for the loss function. In fact, parameter $t_p$ accounts for the degree of robustness that the designer wishes to enforce with respect to the choice of the prior, while parameter $t$ controls robustness to outliers. The $(m,t)$-robust Bayesian learning criterion is a special case of the generalized criterion (\ref{eq:gen_proposed_obj}) for $t_p=1$. Furthermore, the Bayesian learning objective in \cite{li2016renyi,knoblauch2022optimization} is recovered by setting $m=1$ and $t=1$. 

\green{The results presented in previous section extend to the generalized $(m,t)$-robust learning criterion as follows. First, the criterion (\ref{eq:gen_proposed_obj}) can be obtained as a bound on the risk (\ref{ens_risk}), extending Lemma  \ref{lem:rob_pac}, as briefly elaborated in Appendix \ref{appendix_a}.  Furthermore, the influence function analysis developed in Section \ref{sec:if} applies directly also to the generalized criterion (\ref{eq:gen_proposed_obj}), since the derivation therein is only  reliant on the properties of the $t$-logarithm and it does not depend on the choice of the prior regularization. }}

\section{Experiments}
\label{Sec4}
In this section, we first describe a simple regression task with an unimodal likelihood, and then we present results for larger-scale classification and regression tasks. The main aim of these experiments is to provide qualitative and quantitative insights into the performance of $(m,1)$-robust Bayesian learning of \cite{morningstar2020pac} and the proposed robust $(m,t)$-robust Bayesian learning. \green{ All examples are characterized by misspecification and outliers.}

\subsection{Multimodal Regression}
For the first experiment, we modify the regression task studied by \cite{masegosa2019learning} and \cite{morningstar2020pac} in order to capture not only model misspecification but also the presence of outliers as in the contamination model (\ref{cont_mod}). To this end, we assume that the ID distribution $\nu(x)$, with $x=(a,b)$, is given by $\nu(a,b)=p(a)\nu(b|a)$, where the covariate $a$ is uniformly distributed in the interval $[-10.5,10.5]$ -- i.e., $p(a)=1/21$ in this interval and $p(a)=0$ otherwise -- and by a response variable $b$ that is conditionally distributed according to the two-component mixture 
\begin{align}\label{eq:exmix}
    \nu(b|a)&= \mathcal{N}(b|\alpha\mu_a,1),\\
    \alpha&\sim \text{Rademacher},\\
    \mu_a&=7\sin\left(\frac{3a}{4}\right)+\frac{a}{2}.
\end{align}
The OOD component $\xi(x)=\xi(a,b)=p(a)\xi(b)$ also has a uniformly distributed covariate $a$ in the interval  $[-10.5,10.5]$, but, unlike the ID measure, the response variable $b$ is independent of $a$, with a distribution concentrated around $b=0$ as
\begin{align}
    \xi(b)=\mathcal{N}(b|0,0.1).
\end{align}
The parametric model is given by $p(x|\theta)=p(a,b|\theta)=p(a)\mathcal{N}(b|f_\theta(a),1)$, where $f_\theta(a)$ is the output of a three-layer fully connected Bayesian neural network  with 50 neurons and Exponential Linear Unit (ELU) activation functions \cite{clevert2015fast} in the two hidden layers. We consider a Gaussian prior $p(\theta)=\mathcal{N}(0,I)$ over the neural network weights and use a Monte Carlo estimator of the gradient based on the reparametrization trick \cite{kingma2013auto} as in \cite{blundell2015weight}.

Consider first only the effect of misspecification. The parametric model assumes a unimodal likelihood $\mathcal{N}(b|f_\theta(a),1)$ for the response variable, and is consequently misspecified with respect to the ID measure (\ref{eq:exmix}). As a result, the {standard Bayesian learning} leads to a unimodal predictive distribution that approximates the mean value of the response variable, while {$(m,1)$-robust Bayesian learning} can closely reproduce the data distribution \cite{masegosa2019learning,morningstar2020pac}. This is shown in Figure \ref{1a}, which depicts the predictive distribution obtained by minimizing the $m$-free energy criterion $\mathcal{J}^m$ with $m=20$ when using exclusively samples from the ID measure (green dots). In virtue of ensembling, the resulting predictive distribution becomes multimodal, and it is seen to provide a good fit to the data from the ID measure. 

Let us evaluate also the effect of outliers. To this end, in Figure \ref{1b} we consider {$(m,1)$-robust Bayesian learning} and minimize again the $m$-free energy criterion, but this time using a data set contaminated with samples from the OOD component (red points) and with a contamination ratio $\epsilon=0.1$. The predictive distribution is seen to cover not only the ID samples but also the outlying data points. In Figure \ref{1c} and \ref{1d}, we finally plot the predictive distributions obtained by {$(m,t)$-robust Bayesian learning} with $m=20$, when setting  $t=\{0.9,0.8\}$, respectively. The proposed approach  is able to mitigate the effect of the outlying component for $t=0.9$, and, for $t=0.8$, it almost completely suppresses it. As a result, the proposed energy criterion produces predictive distributions that match more closely the ID measure. This qualitative behavior is quantified in Table \ref{tab:first_exp}, where we report the total variation distance from the ID measure for the setting and predictors considered in Figure 4.

\subsection{\blue{MNIST and CIFAR-10 Classification Tasks}}\label{sec:class}

\begin{figure*}
    \centering
    \subcaptionbox{\label{5.1a} MNIST data set}
        [0.345\textwidth]{ \includegraphics[width=0.35\textwidth]{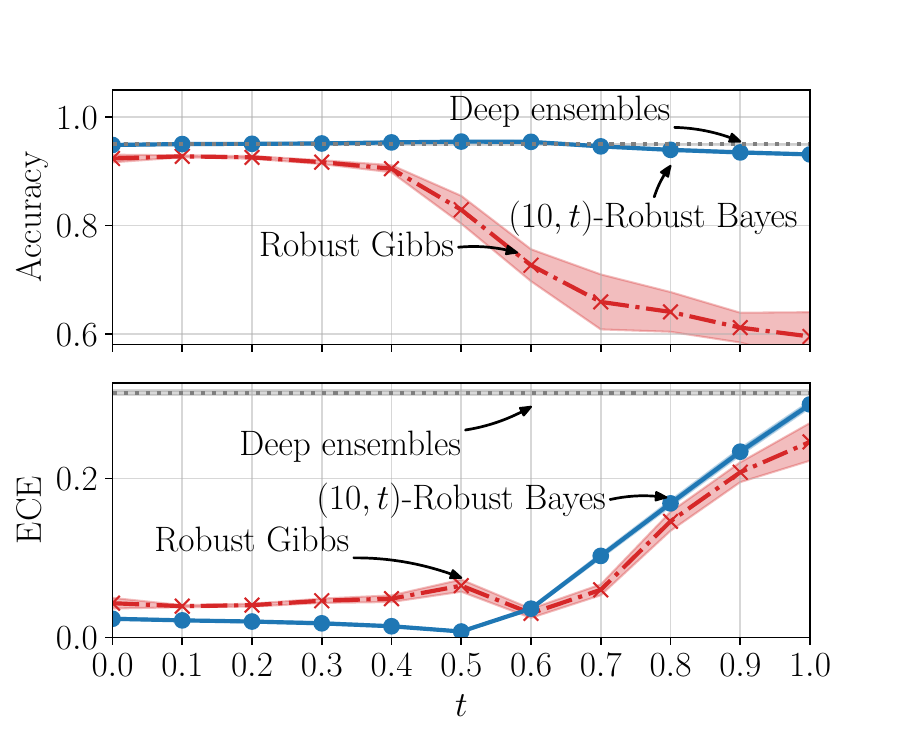}}
        \hspace{-1.75em}
   \subcaptionbox{\label{5.1b} Extended MNIST data set}
        [0.345\textwidth]{ \includegraphics[width=0.35\textwidth]{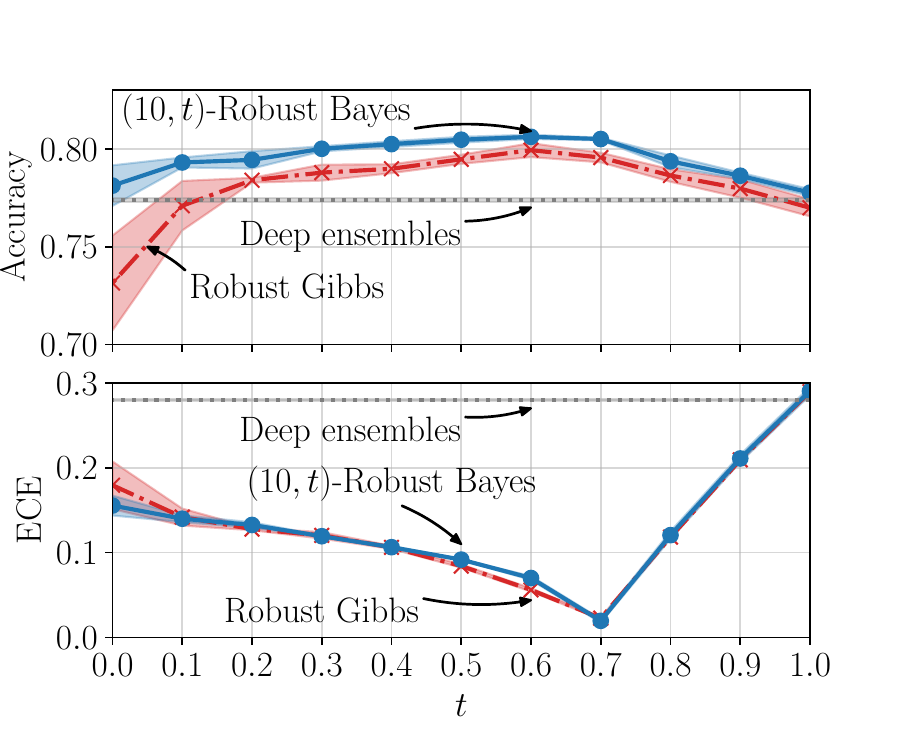}}
         \hspace{-1.75em}
    \subcaptionbox{\label{5.1c} CIFAR-10 data set}
        [0.345\textwidth]{ \includegraphics[width=0.35\textwidth]{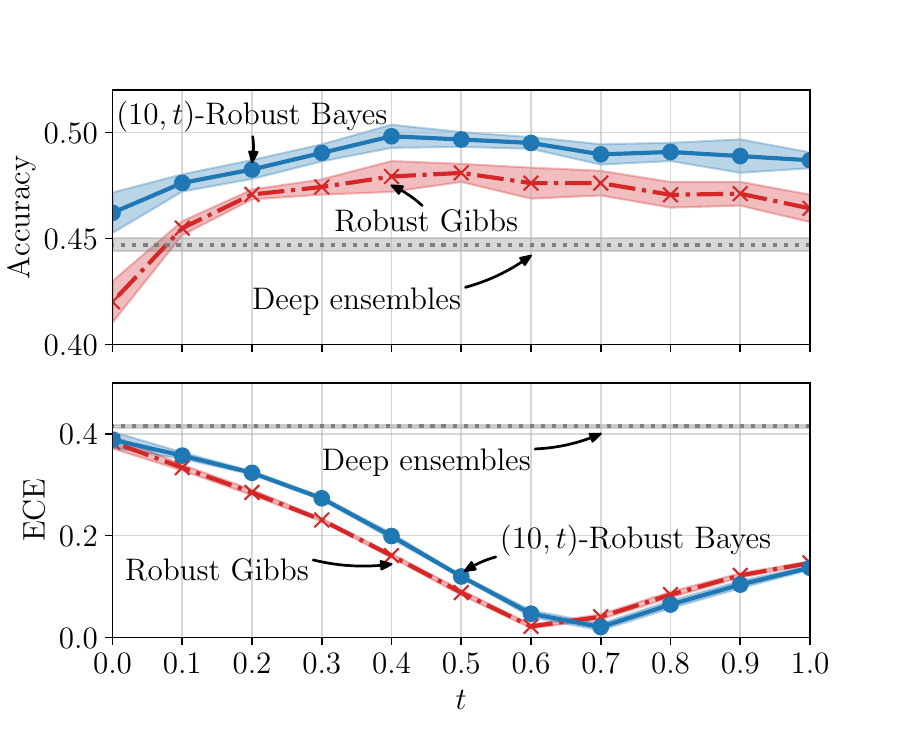}}
    \caption{\blue{Test accuracy (top) and expected calibration error (ECE) (bottom) as a function of $t$ under the contamination ratio $\epsilon=0.3$ for: \emph{(i)} deep ensembles \cite{lakshminarayanan2017simple}; \emph{(ii)} robust Gibbs predictor, which minimizes the free energy criterion $\mathcal{J}_t^1$ \cite{amid2019robust}; and \emph{(iii)} $(m,t)$-robust Bayesian learning, which minimizes the free energy criterion $\mathcal{J}_t^{10}$.}}
    \label{fig:class_acc_and_ece}
\end{figure*}

\begin{figure}
    \centering
    \hspace{-1em}
    \includegraphics[width=0.5\textwidth]{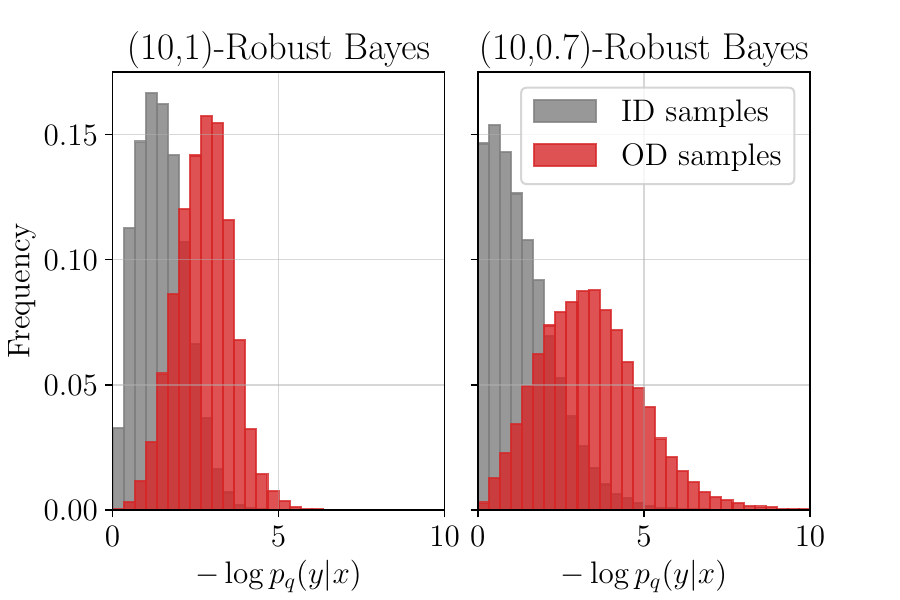}
    \caption{\blue{Distribution of the negative $\log$-likelihood of ID and OD training data samples for an ensemble model minimizing (on the left) the $\log$-loss based criterion $\mathcal{J}_{1}^{10}$, and (on the right) the proposed robust objective $\mathcal{J}_{0.7}^{10}$ based on the $\log_t$-loss with $t=0.7$. }}
    \label{fig:Loss_dist}
\end{figure}

\blue{
We now address the problem of training Bayesian neural network classifiers in the presence of misspecification and outliers. We consider three different experimental setups entailing distinct data sets and model architectures:
\begin{itemize}
    \item Classification of MNIST digits \cite{lecun1998mnist} based on a fully connected neural network comprising a single hidden layer with $25$ neurons.
    \item Classification of Extended MNIST characters and digits \cite{cohen2017emnist} based on a fully connected neural network with two hidden layers with $25$ neurons each.
    \item Classification of CIFAR-10 \cite{krizhevsky2009learning} images using a convolutional neural network (CNN) with two convolutional layers, the first with $8$ filters of size $3\times3$ and the second with $4$ filters of size $2\times2$, followed by a hidden layer with $25$ neurons each.
\end{itemize}
All hidden units use ELU activations \cite{clevert2015fast} except the last, classifying, layer that implements the standard softmax function.
Model misspecification is enforced by adopting neural network architectures with small capacity.
As in \cite{amid2019robust}, outliers are obtained by randomly modifying the labels for fraction $\epsilon$ of the data points in the training set. Additional details for the experiments can be found in the supplementary material.}

\blue{
We measure the accuracy of the trained models, as well as their calibration performance. Calibration refers to the capacity of a model to quantify uncertainty (see, e.g., \cite{lakshminarayanan2017simple}). We specifically adopt the expected calibration error (ECE) \cite{guo2017calibration}, a standard metric that compares model confidence to actual test accuracy (see supplementary material for the exact definition). 
We train the classifiers using corrupted data sets with a contamination ratio $\epsilon=0.3$, and then we evaluate their accuracy and ECE as a function of $t \in [0, 1]$ based on a clean ($\epsilon=0$) holdout data set. We compare the performance of $(m,t)$-robust Bayesian learning based on the minimization of the robust $m$-free energy $\mathcal{J}_t^{m}$, with $m=10$, to: \emph{(i)} \emph{deep ensembles}  \cite{lakshminarayanan2017simple}, also with $10$ models in the ensembles; and \emph{(ii)} the robust Gibbs predictor of  \cite{amid2019robust}, which optimizes over a single predictor (not an ensemble) by minimizing the free energy metric $\mathcal{J}_t^{1}$. The inverse temperature parameter $\beta$ is set to $0.1$ in the $(m,t)$-robust Bayesian and the Gibbs predictor objectives.}

\blue{
In Figure \ref{fig:class_acc_and_ece} we report the performance metrics attained by the trained models in the three different setups listed above. From the top panels we conclude that  $(m,t)$-robust Bayesian learning is able to mitigate model misspecification by improving the final accuracy as compared to the robust Gibbs predictor and the deep ensemble models. Furthermore, the use of the robust loss for a properly chosen value of $t$ leads to a reduction of the detrimental effect of outliers and to an increase in the model accuracy performance as compared to the standard $\log$-loss ($t=1$).  }
\blue{
In terms of calibration performance, the lower panels demonstrate the capacity of robust ensemble predictors with $t<1$ to drastically reduce the ECE as compared to deep ensembles. In this regard, it is also observed that the accuracy and ECE performance levels depend on the choice of parameter $t$. In practice, the selection of $t$ may be addressed using validation or meta-learning methods in a manner akin to \cite{zhang2021meta}. Additional results on calibration in the form of reliability diagrams \cite{degroot1983comparison} can be found in supplementary material. }

\blue{
As shown shown theoretically in Section \ref{sec:if}, the effect of the $\log_t$-loss is to reduce the influence of outliers during training for $t<1$. We empirically investigate the effect of the robust loss in Figure \ref{fig:Loss_dist}, in which we compare the distribution of the negative $\log$-likelihood for ID and OD training data samples. We focus on the CIFAR-10 data set, and we compare the histogram of the negative $\log$-likelihood under a CNN model trained based on the $m$-free energy $\mathcal{J}_1^{m}$, with $m=10$ and standard logarithmic loss, and a CNN minimizing the proposed robust $m$-free energy $\mathcal{J}_t^{m}$, with $m=10$ and $t=0.7$. The $(m,1)$-robust Bayesian based on the standard $\log$-loss tries to fit both ID and OD samples and, as a result, the two components have similar likelihoods. In contrast, $(m,t)$-robust Bayesian learning is able to downweight the influence of outliers and to better fit the ID component. }

\subsection{\blue{California Housing Regression Task}}\label{sec:class}
\begin{figure*}
    \centering
    \hspace{-1em}
    \includegraphics[width=1\textwidth]{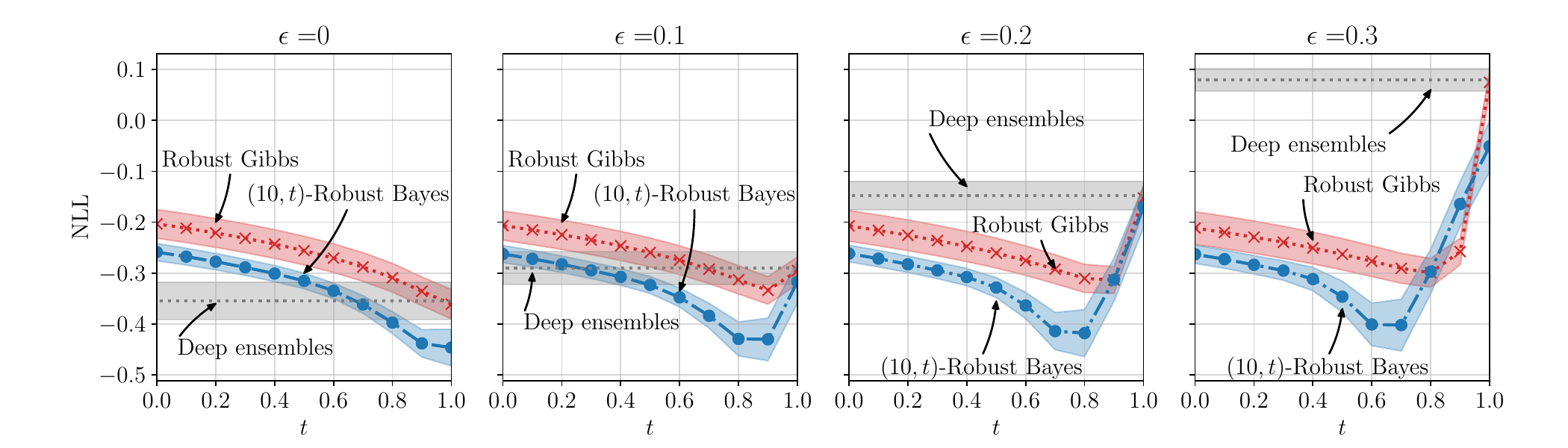}
    \caption{\blue{ Negative $\log$-likelihood computed on a uncorrupted data set for: \emph{(i)} deep ensembles \cite{lakshminarayanan2017simple}; \emph{(ii)} robust Gibbs predictor, which minimizes $\mathcal{J}_t^1$ \cite{amid2019robust}; and \emph{(iii)} the $(m,t)$-robust Bayesian learning, which minimizes $\mathcal{J}_t^{10}$. The models are trained on $\epsilon$-contaminated data set for $\epsilon\in\{0,0.1,0.2,0.3\}$ }}
    \label{fig:cali_house}
\end{figure*}
\blue{
We consider the problem of training a robust regressor based on training data sets corrupted by outliers and in the presence of model misspecification. We consider the California housing dataset, which is characterized by response variables $y$ normalized in the $[0,1]$ interval, and we fix a unimodal likelihood $p(y|x,\theta)=\mathcal{N}(y|f_\theta(x),0.1)$, where $f_\theta(x)$ is the output of a three-layer neural network with hidden layers comprising 10 units with ELU activation functions \cite{clevert2015fast}. We consider a Gaussian prior  $p(\theta)=\mathcal{N}(\theta|0,I)$. The model class is misspecified since  the response variable is bounded and hence not Gaussian. Outliers are modeled by replacing the label of fraction $\epsilon$ of the training sample with random labels picked uniformly at random within the $[0,1]$ interval.}

\blue{
We consider training based on data sets with different contamination ratios $\epsilon\in\{0,0.1,0.2,0.3\}$, and measure the trained model ability to approximate the ID data by computing the negative $\log$-likelihood on a clean holdout data set ($\epsilon=0$).
As in the previous subsection, we compare models trained using $(m,t)$-robust Bayesian learning, with $m=5$, to: \emph{(i)} \emph{deep ensembles} \cite{lakshminarayanan2017simple}, also with $5$ models in the ensembles; and \emph{(ii)} the robust Gibbs predictor of \cite{amid2019robust} minimizing the free energy metric $\mathcal{J}_t^{1}$. The inverse temperature parameter $\beta$ is set to 0.1 in the $(m,t)$-robust Bayesian and the Gibbs predictor objectives.}

\blue{
In Figure \ref{fig:cali_house} we report the negative $\log$-likelihood of an uncontaminated data set for models trained according to the different learning criteria. The leftmost panel ($\epsilon=0$) corresponds to training based on an uncontaminated data set. For this case, the best performance is obtained for $t=1$ -- an expected result due to the absence of outliers -- and the proposed criterion outperforms both the Gibbs predictor and deep ensembles, as it is capable of counteracting misspecification by the means of ensembling.
In the remaining panels, training is performed based on $\epsilon$-contaminated data sets, with the contamination $\epsilon$ increasing from left to right. In these cases, learning criteria based on robust losses are able to retain similar performance to the uncontaminated case for suitable chosen values of $t$. Furthermore, the optimal value of $t$ is observed to increase with the fraction of outliers in the training data set.
}

\subsection{\green{Robustness to Prior Misspecification}}\label{sec:pr_miss}

\begin{figure}
    \centering
    \hspace{-1em}
    \includegraphics[width=0.5\textwidth]{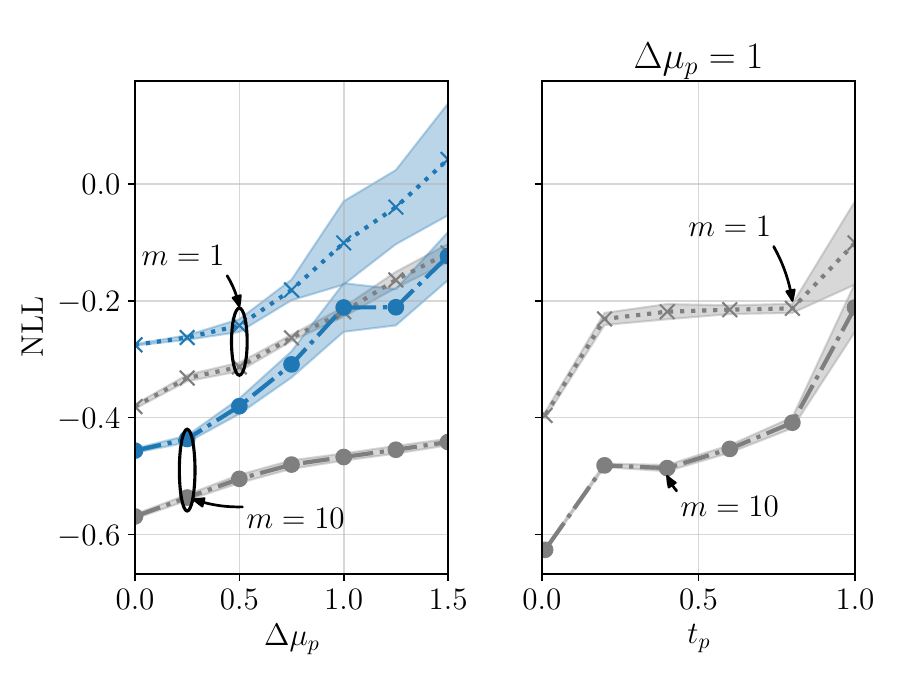}
    \caption{\green{Negative log-likelihood obtained minimizing the free energy with the Rényi entropy and different different values of $t_p$. In the left panel, we consider the generalized $(m,t)$-robust Bayesian learning for $t=1$ and $m\in\{1,10\}$ using the standard KL regularizer (in blue) and the Rényi regularizer for $t_p=0.5$ (in gray). In the right panel we fix the parameter $\Delta\mu_p=1$ and evaluate the performance as a function of $t_p$.}}
    \label{fig:pr_miss}
\end{figure}

{\color{black}
We finally turn to exploring the robustness of the generalized $(m,t)$-robust criterion (\ref{eq:gen_proposed_obj}) with respect to the choice of the prior distribution. To this end, we consider the same regression problem and likelihood model of the previous subsection, but we allow for a Gaussian prior distribution $p(\theta|\Delta\mu_p)=\mathcal{N}(\theta|\Delta\mu_pI,0.1I)$ with a generally non-zero mean $\Delta\mu_p$. In Bayesian neural network training, it is customary to set $\Delta\mu_p=0$, favoring posteriors with a small expected norm, which are expected to generalize better \cite{dusenberry2020efficient,hernandez2015probabilistic}. In order to study the impact of misspecification, similarly to \cite{knoblauch2019frequentist}, we evaluate the performance obtained with different prior regularizers  when choosing a non-zero prior mean $\Delta\mu_p$. Non-zero values of the prior may be considered to be misspecified as they do not comply with the actual expectation on the best model parameters for this problem.

In the leftmost panel of Figure \ref{fig:pr_miss}, we show the negative log-likelihood obtained by $(m,t)$-robust Bayesian learning for $t=1$ and $m\in\{1,10\}$, which uses the standard KL regularizer (in blue), as well as by  generalized robust  Bayesian learning  with the Rényi regularizer with $t_p=0.5$ (in gray). The advantage of the generalized approach is particularly apparent for $m=10$, in which case generalized $(m,t)$-robust Bayesian learning with $t_p=0.5$ shows a more graceful performance degradation for increasing values of $\Delta\mu_p$. }
{\color{black}

To further elaborate on the role of the choice of the parameter $t_p$, in the rightmost panel, we fix the prior parameter as $\Delta{\mu_p}=1$, and plot the negative log-likelihood as a function of $t_p$ for $m=1$ and $m=10$. In both cases, we find that the robustness to a misspecified prior increases as $t_p$ decreases, demonstrating the advantages of the generalized robust Bayesian learning framework.
}

\section{Conclusion}
In this work, we addressed the problem of training ensemble models under model misspecification and in the presence of outliers. We proposed the $(m,t)$-robust Bayesian learning framework that leverages generalized logarithm score functions in combination with multi-sample bounds, with the goal of deriving posteriors that are able to take advantage of ensembling, while at the same time being robust with respect to outliers. The proposed learning framework is shown to lead to predictive distributions characterized by better generalization capabilities and calibration performance in scenarios in which the standard Bayesian posterior fails.

\blue{The proposed robust Bayesian learning framework can find application to learning scenarios that can benefit from uncertainty quantification in their decision making processes and are characterized by the presence of outliers and model misspecification. Examples include inference in wireless communication systems \cite{zecchin2022robust}, medical imaging \cite{liu2020time} and text sentiment analysis \cite{onan2017hybrid,onan2018biomedical}.}

\blue{
We conclude by suggesting a number of directions for future research. The $(m,t)$-robust Bayesian learning has been shown to lead to the largest performance gains for properly chosen values of $t$. The optimal values of $t$ depend on the particular task at hand, and deriving rules to automate the tuning of these parameters represents a practical and important research question. Furthermore, $(m,t)$-robust Bayesian learning can be extended to reinforcement learning, as well as to meta-learning, for which Bayesian methods have recently been investigated (see, e.g., \cite{yoon2018bayesian,jose2022information} and references therein). }

\bibliographystyle{IEEEtran}
\bibliography{uai2022-template}

\appendix

\subsection{Proofs}
\label{appendix_a}
\begin{lemma*}With probability $1-\sigma$, with $\sigma\in(0,1)$, with respect to the random sampling of the data set $\mathcal{D}$, for all distributions $q(\theta)$ that are absolutely continuous with respect the prior $p(\theta)$, the following bound on the risk ($\ref{ens_risk}$) of the ensemble model holds
\begin{align}
    \mathcal{R}_t(q)\leq&\mathcal{J}^m_t(q)+\psi(\tilde{\nu},n,m,\beta,p,\sigma),
\end{align}
where 
\begin{align}
   \psi(\tilde{\nu},n,m,\beta,p,\sigma):=\frac{1}{\beta}\left(\log\mathbb{E}_{\mathcal{D},p(\theta)}\left[e^{\beta\Delta_{m,n}}\right]-\log\sigma\right)
\end{align}
and
\begin{align}
   \Delta_{m,n}:=&\frac{1}{n}\sum_{x\in \mathcal{D}}\log_t\hspace{-0.2em}\mathbb{E}_{j\sim U[1:m]}p(x|\theta_j)\nonumber\\
   &-\mathbb{E}_{\tilde{\nu}(x)}\left[\log_t\hspace{-0.2em}\mathbb{E}_{j\sim U[1:m]}p(x|\theta_j)\right].
\end{align}
Furthermore, the risk with respect to the ID measure $\nu(x)$ can be bounded as 
\begin{align}
    \mathbb{E}_{\nu(x)}[\mathcal{R}_t(q,x)]\leq&\frac{1}{1-\epsilon}\left( \mathcal{J}^m_t(q)+\psi(\tilde{\nu},n,m,\beta,p,\sigma)\right)\nonumber\\
    &+\frac{\epsilon(C^{1-t}-1)}{(1-\epsilon)(1-t)},
\end{align}
if the contamination ratio satisfies the inequality $\epsilon<1$.
\end{lemma*}

\textbf{Proof:}
The proof follows in a manner similar to \cite{morningstar2020pac}. For a data set size $n$, and for an ensemble of models $\Theta=\{\theta\}^m_{i=1}$, we define the quantity 
\begin{align}
\Delta_{m,n}(\Theta,\mathcal{D}):=&\frac{1}{n}\sum_{x\in \mathcal{D}}\log_t\hspace{-0.2em}\mathbb{E}_{j\sim U[1:m]}p(x|\theta_j)\nonumber\\
&-\frac{1}{n}\sum_{x\in \mathcal{D}}\mathbb{E}_{\tilde{\nu}(x)}\left[\log_t\hspace{-0.2em}\mathbb{E}_{j\sim U[1:m]}p(x|\theta_j)\right].\label{change_measure}
\end{align}
From the compression lemma \cite{banerjee2006bayesian}, we have that for any distribution $q(\theta)$ which is absolutely continuous with respect to the prior $p(\theta)$, and for any $\beta<0$, the following holds
\begin{align}
   \mathbb{E}_{q(\theta)^{\otimes m}}\left[\beta\Delta_{m,n}\right]\leq& D_1(q(\theta)^{\otimes m}||p(\theta)^{\otimes m})\nonumber\\
   &+\log \mathbb{E}_{p(\theta)^{\otimes m}}\left[e^{\beta\Delta_{m,n}}\right]\\
   =& mD_1(q(\theta)||p(\theta))\nonumber\\
   &+\log \mathbb{E}_{p(\theta)^{\otimes m}}\left[e^{\beta\Delta_{m,n}}\right],\label{proof_eq1}
\end{align}
where we have used the simplified notation $\Delta_{m,n}=\Delta_{m,n}(\Theta,\mathcal{D})$, and the equality follows from the basic properties of the KL divergence.

A direct application of Markov's inequality is then used to bound the last term of (\ref{proof_eq1}) with high probability.
Namely, with probability greater then $1-\sigma$ with respect to the random drawn of the data set $\mathcal{D}\sim \tilde{\nu}(x)^{\otimes n}$, the following holds 
\begin{align}
   \mathbb{E}_{p(\theta)^{\otimes m}}\left[e^{\Delta_{m,n}}\right]\leq  \frac{\mathbb{E}_{\tilde{\nu}(x)^{\otimes n},p(\theta)^{\otimes m}}\left[e^{\Delta_{m,n}}\right]}{\sigma},
\end{align}
or, equivalently,
\begin{align}
  \log\mathbb{E}_{p(\theta)^{\otimes m}}\left[\hspace{-0.1em}e^{\Delta_{m,n}}\hspace{-0.1em}\right]\hspace{-0.1em}\leq \hspace{-0.1em} \log\mathbb{E}_{\tilde{\nu}(x)^{\otimes n},p(\theta)^{\otimes m}}\left[\hspace{-0.1em}e^{\Delta_{m,n}}\hspace{-0.1em}\right]\hspace{-0.1em}-\hspace{-0.1em}\log\sigma. \label{proof_eq2}
\end{align}
Combining (\ref{proof_eq1}) with (\ref{proof_eq2}), the following upper bound on the predictive risk holds with probability $1-\sigma$
\begin{align}
\mathcal{R}_t(q)\leq&\mathbb{E}_{\tilde{\nu}(x),q(\theta)^{\otimes m}}\left[-\log_t\hspace{-0.2em}\mathbb{E}_{j\sim U[1:m]}p(x|\theta_j)\right]\\
\leq &\mathbb{E}_{q(\theta)^{\otimes m}}\left[\frac{1}{n}\sum_{x\in \mathcal{D}}\log_t\hspace{-0.2em}\mathbb{E}_{j\sim U[1:m]}p(x|\theta_j)\right]\nonumber\\
&+\frac{m}{\beta}D_1(q(\theta)||p(\theta))\nonumber\\
&+\frac{\log\mathbb{E}_{\tilde{\nu}(x)^{\otimes n}}\mathbb{E}_{p(\theta)^{\otimes m}}\left[e^{\Delta_{m,n}}\right]-\log\sigma}{\beta}.
\end{align}
Finally, the result above can be translated to a guarantee with respect to the ID measure $\nu(x)=\frac{\tilde{\nu}(x)}{1-\epsilon}-\frac{\epsilon}{1-\epsilon}\xi(x)$ via the sequence of inequalities
\begin{align}
    \mathbb{E}_{\nu(x),q(\theta)^{\otimes m}}&\left[-\log_t\hspace{-0.2em}\mathbb{E}_{j\sim U[1:m]}p(x|\theta_j)\right]=\nonumber\\
    &=\frac{\mathbb{E}_{\tilde{\nu}(x),q(\theta)^{\otimes m}}\left[-\log_t\hspace{-0.2em}\mathbb{E}_{j\sim U[1:m]}p(x|\theta_j)\right]}{1-\epsilon}\nonumber\\
    &+\epsilon\frac{\mathbb{E}_{\epsilon(x),q(\theta)^{\otimes m}}\left[-\log_t\hspace{-0.2em}\mathbb{E}_{j\sim U[1:m]}p(x|\theta_j)\right]}{1-\epsilon}\\
    &\leq\frac{\mathbb{E}_{\tilde{\nu}(x),q(\theta)^{\otimes m}}\left[-\log_t\hspace{-0.2em}\mathbb{E}_{j\sim U[1:m]}p(x|\theta_j)\right]}{1-\epsilon}\nonumber\\
    &+\epsilon\frac{\left(C^{1-t}-1\right)}{(1-\epsilon)(1-t)},
\end{align}
where the last inequality follows by having assumed the probabilistic model being uniformly upper bounded by $C$ (Assumption \ref{ass:mis}).

\hfill $\blacksquare$

\green{The above result can readily be extended to generalized robust Bayesian learning by applying the change of measure inequality presented in \cite{begin2016pac}, namely
\begin{align}
    \frac{t_p}{1-t_p}\log\mathbb{E}_{q(\theta)^{\otimes m}}\phi(\Theta)\leq&  D^R_{t_p}(q(\theta)^{\otimes m}||p(\theta)^{\otimes m})\\
    &+\log \mathbb{E}_{p(\theta)^{\otimes m}}\left[\phi(\Theta)^{ \frac{t_p}{1-t_p}}\right],
\end{align}
with $\phi(\Theta)$ being the function 
\begin{align}
    \phi(\Theta):=e^{\frac{t_p}{1-t_p}\Delta_{m,n}(\Theta,\mathcal{D})},
\end{align}
and by exploiting the tensorization of the Rényi divergence 
\begin{align}
    D^R_{t_p}(q(\theta)^{\otimes m}||p(\theta)^{\otimes m})=m D^R_{t_p}(q(\theta)||p(\theta)).
\end{align}}

Finally, with regard to the comparison between the PAC$^m$ bound in Theorem 1 in \cite{morningstar2020pac} and the guarantee with respect to the ID measure, we observe that it is not in general possible to translate a guarantee on the $\log_t$-risk to one on the $\log$-risk. This can be illustrated by the following counter-example. Consider the following discrete target distribution parametrized by integer $k$, which defines the size of its support, as
 \begin{align}
    \nu_k(x)=
\begin{cases}
1-\frac{1}{k},\ &\text{ for } x=0\\
\frac{1}{k}2^{-k^2},\ &\text{ for } x=1,\dots,2^{k^2},\\
\end{cases}
 \end{align}
and the optimization of the $\log_t$-loss over a predictive distribution $p(x)$. The following limit holds
 \begin{align}
     \lim_{k\to\infty}\min_{p}\mathbb{E}_{\nu_k(x)}[\log_tp(x)]=\begin{cases}
0,\ &\text{ for } t\in[0,1)\\
\infty,&\text{ for } t=1\\
\end{cases},
\end{align}
and therefore that an ensemble optimized for a value of $t$ in the range $[0,1)$ can incur in an unboundedly large loss when scored using the $\log$-loss.

\section{Proof of Theorem \ref{thm1}}
\begin{theorem*}
The minimizer of the robust $m$-free energy objective 
\begin{align}
    \mathcal{J}^m_t(q):=\frac{1}{n}\sum_{x\in\mathcal{D}}\hat{\mathcal{R}}^m_t(q,x)+\frac{m}{\beta} D_1(q(\theta)||p(\theta)).
\end{align}
is the fixed point of the operator
\begin{align}
T(q)\hspace{-0.2em}:=\hspace{-0.2em}p(\theta_j)\exp\left(\hspace{-0.2em}\beta\hspace{-0.2em}\sum_{x\in\mathcal{D}}\hspace{-0.1em}\mathbb{E}_{\{\theta_i\}_{i\neq j}}\hspace{-0.3em}\left[\log_t\hspace{-0.2em}\left(\frac{\sum^m_{i=1} p_{\theta_i}(x)}{m}\right)\right]\hspace{-0.2em}\right)
\end{align}
where the average in (\ref{minimizer_rob}) is taken with respect to the i.i.d. random vectors $ \{\theta_i\}_{i\neq j}\sim q(\theta)^{\otimes m-1}$.
\end{theorem*}
\textbf{Proof:} The functional derivative of the multi-sample risk is instrumental to computation of the minimizer of the robust $m$-free energy objective (\ref{eq:proposed_obj}). This is given as
 \begin{align}
     &\hspace{0.2em}\frac{d\hat{\mathcal{R}}^m_t(q,x)}{dq}=\nonumber\\
     &=\frac{d}{dq}\mathbb{E}_{\theta_1,\dots,\theta_m\sim q(\theta)^{\otimes m}}\left[-\log_t\hspace{-0.2em}\mathbb{E}_{j\sim U[1:m]}p(x|\theta_j)\right]\\
     &=-\frac{d}{dq}\int_{\Theta^m}\log_t\hspace{-0.2em}\mathbb{E}_{j\sim U[1:m]}p(x|\theta_j)\prod^m_{i=1}q(\theta_i)d\theta_i\\
     &\numeq{a}-\sum^m_{k=1}\int_{\Theta^{m-1}}\log_t\hspace{-0.2em}\mathbb{E}_{j\sim U[1:m]}p(x|\theta_j)\prod_{i\neq k}q(\theta_i)d\theta_i\\
     &\numeq{b}-m\int_{\Theta^{m-1}}\log_t\hspace{-0.2em}\mathbb{E}_{j\sim U[1:m]}p(x|\theta_j)\prod^{m-1}_{i=1 }q(\theta_i)d\theta_i,\\
     &=-m\mathbb{E}_{\theta_1,\dots,\theta_{m-1}\sim q(\theta)^{\otimes {m-1}}}\left[\log_t\hspace{-0.2em}\mathbb{E}_{j\sim U[1:m]}p(x|\theta_j)\right],
\end{align}
where $(a)$ follows from the derivative of a nonlocal functional of $m$ functions, and $(b)$ holds since the integrand is invariant under the permutation of $\{\theta_i\}_{i\neq k}$.

The functional derivative of the robust $m$-free energy then follows as 
\begin{align}
     &\frac{d\mathcal{J}^m_t(q)}{dq}\hspace{-0.2em}=\\
     &=\frac{d\hat{\mathcal{R}}^m_t(q,x)}{dq}+\frac{m}{\beta}\frac{dD_1(q(\theta)||p(\theta)}{dq}\\
     &=-m\mathbb{E}_{\theta_1,\dots,\theta_{m-1}\sim q(\theta)^{\otimes {m-1}}}\left[\log_t\hspace{-0.2em}\mathbb{E}_{j\sim U[1:m]}p(x|\theta_j)\right]\\
     &+\frac{m}{\beta}\left(1+\log(q(\theta))-\log(p(\theta))\right).
\end{align}
Imposing the functional derivative equals to zero function it follows that the optimized posterior must satisfy
\begin{align}
     q(\theta_m)=&p(\theta_m)\cdot\\
     &\hspace{-3em}\cdot\exp{\beta\mathbb{E}_{\theta_1,\dots,\theta_{m-1}\sim q(\theta)^{\otimes {m-1}}}\left[\log_t\hspace{-0.2em}\mathbb{E}_{j\sim U[1:m]}p(x|\theta_j)\right]}.
\end{align}
\hfill $\blacksquare$

\section{Proof of Theorem \ref{IF_th}}
\begin{theorem*}
The influence function of the robust $m$-free energy objective (\ref{PAC-t contaminated}) is
\begin{align}
\label{app:IF}
IF^m_t\hspace{-0.1em}(\hspace{-0.1em}z,\phi,P^n\hspace{-0.1em})&\hspace{-0.2em}=\hspace{-0.2em}-\hspace{-0.2em}\left[\hspace{-0.1em}\pdv[2]{\mathcal{J}^m_t(\gamma,\phi)}{\phi}\hspace{-0.1em}\right]^{\hspace{-0.1em}-1}\hspace{-1.2em}\times\hspace{-0.2em}\pdv[2]{\mathcal{J}^m_t(\gamma,\phi)}{\gamma}{\phi}\hspace{-0.1em}\Bigg|_{\substack{\gamma=0\hfill\\\phi=\phi^{m*}_t\hspace{-0.2em}(0)}}\hspace{-0.4em},
\end{align}
where
\begin{align}
\pdv[2]{\mathcal{J}^m_t(\gamma,\phi)}{\phi}\hspace{-0.2em}=&\mathbb{E}_{P^n_{\gamma,z}(x)}\pdv[2]{}{\phi}\left[\hat{\mathcal{R}}^m_t(q_{\phi},x)\right]\nonumber\\
&+\pdv[2]{}{\phi}\left[\frac{m}{\beta} KL(q_{\phi}(\theta)||p(\theta))\right]
\end{align}
and
\begin{align}
\pdv[2]{\mathcal{J}^m_t(\gamma,\phi)}{\gamma}{\phi}\hspace{-0.2em}=\hspace{-0.2em}\pdv{}{\phi}\left[\mathbb{E}_{ P^n(x)}\left[\hat{\mathcal{R}}^m_t(q_{\phi},x)\right]\hspace{-0.2em}-\hspace{-0.2em}\hat{\mathcal{R}}^m_t(q_{\phi},z)\right].
\end{align}
\end{theorem*}
The proof of Theorem \ref{IF_th} directly follows from the Cauchy implicit function theorem stated below.
\begin{thm}[Cauchy implicit function theorem]
Given a continuously differentiable function $F:\mathbb{R}^n\times \mathbb{R}^m \to \mathbb{R}^m$, with domain coordinates $(x,y)$, and a point $(x^*,y^*)\in \mathbb{R}^n\times \mathbb{R}^m$ such that $F(x^*,y^*)=0$, if the Jacobian $J_{F,y}(x^*,y^*)=\left[\pdv{F_1(x^*,y^*)}{y_1}, \dots, \pdv{F_m(x^*,y^*)}{y_m}\right]$ is invertible, then there exists an open set $U$ that contains $x^*$ and a function $g:U\to Y$ such that $g(x^*)=y^*$ and $F(x,g(x))=0$, $\forall x \in U$. Moreover the partial derivative of $g(x)$ in $U$ are given by
\begin{align}
{\pdv{g}{x_i}}(x)=-\left[J_{F,y}(x,g(x))\right]^{-1}\left[{\pdv{F}{x_i}} (x,g(x))\right]
\label{IFT}
\end{align}
\end{thm}

\textbf{Proof:} Replacing $F(x,y)$ with $\pdv{\mathcal{J}^m_t(\gamma,\phi)}{\phi}$ and $g(x)$ with $\phi^{m*}_t(\gamma)$ and accordingly rewriting ($\ref{IFT}$), we obtain
\begin{align}
    \frac{d\phi^{m*}_t(\gamma)}{d\gamma}=\hspace{-0.3em}-\hspace{-0.3em}\left[\hspace{-0.1em}\pdv[2]{\mathcal{J}^m_t(\gamma,\phi^{m*}_t(\gamma))}{\phi}\hspace{-0.2em}\right]^{\hspace{-0.1em}-1}\hspace{-0.8em}\times\hspace{-0.2em}\pdv[2]{\mathcal{J}^m_t(\gamma,\phi^{m*}_t(\gamma))}{\gamma}{\phi}.
    \label{IF_not_eval}
\end{align}
The influence function (\ref{app:IF}) is then obtained evaluating (\ref{IF_not_eval}) at $\gamma=0$.

\hfill $\blacksquare$

\section{Simulation Details}
\begin{figure*}
    \centering
    \subcaptionbox{\label{app:1a}$m=1$}
        [0.24\textwidth]{ \includegraphics[width=0.24\textwidth]{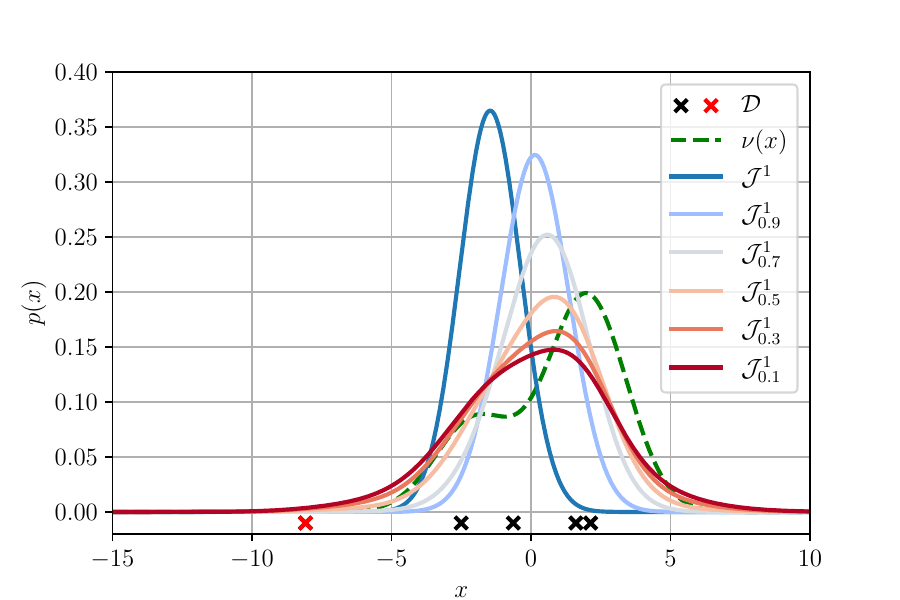}}
   \subcaptionbox{\label{app:1b}$m=2$}
        [0.24\textwidth]{ \includegraphics[width=0.24\textwidth]{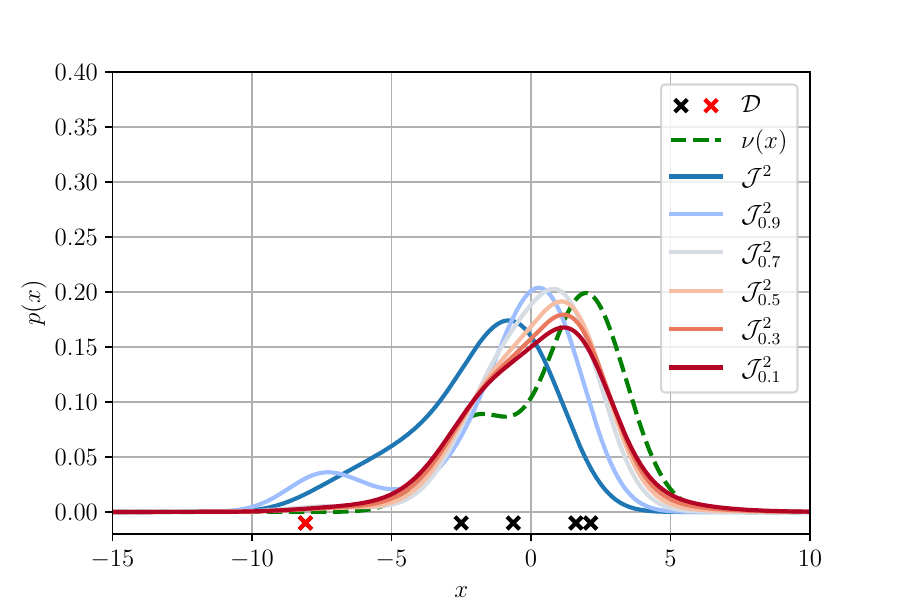}}
    \subcaptionbox{\label{app:1c}$m=5$}
        [0.24\textwidth]{ \includegraphics[width=0.24\textwidth]{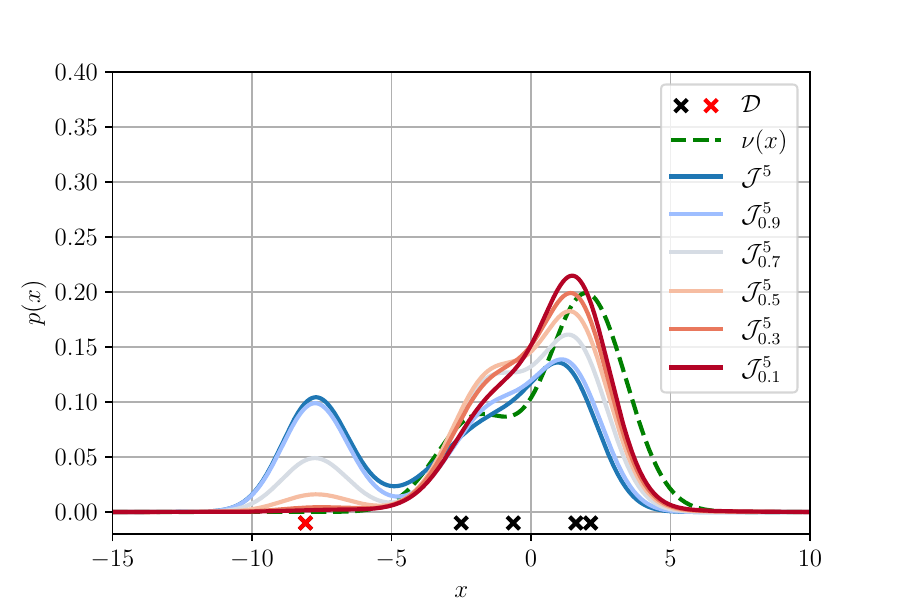}}
    \subcaptionbox{\label{app:1d}$m=20$}
        [0.24\textwidth]{ \includegraphics[width=0.24\textwidth]{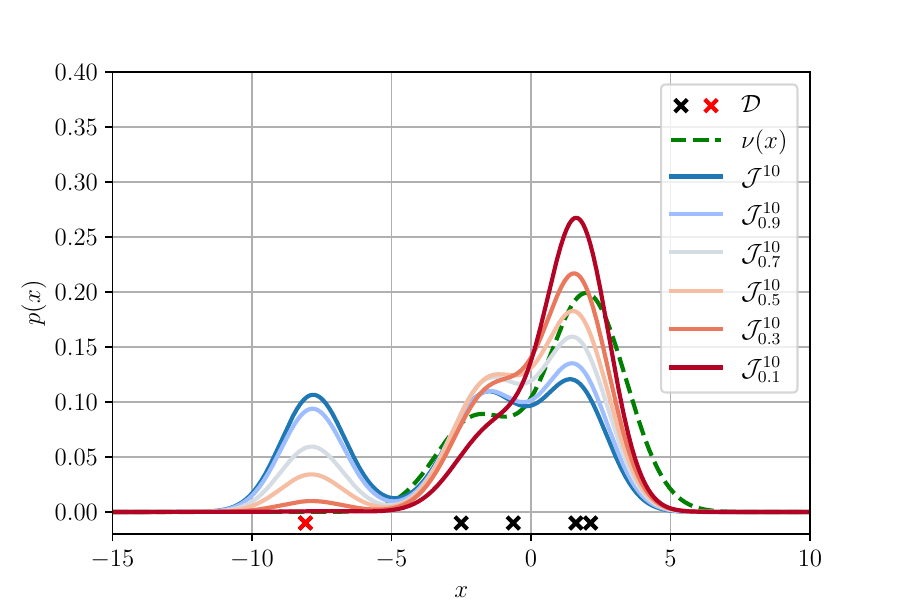}}
    \caption{Ensemble predictive distribution obtained minimizing different free energy criteria and different values of $m$. The samples from the ID measure are represented as green dots, while data points sampled from the OOD component are in red. The optimized predictive distributions. The predictive distribution obtained minimizing the standard $m$-free energy is denoted by $\mathcal{J}^{m}$, while the predictive distribution yielded by the minimization of the robust $m$-free energy are denoted by $\mathcal{J}^{m}_{0.9},\mathcal{J}^{m}_{0.7},\mathcal{J}^{m}_{0.5}, \mathcal{J}^{m}_{0.3}$  and $\mathcal{J}^{m}_{0.1}$ for $t=\{1,0.9,0.7,0.5,0.3,0.1\}$ respectively. }
    \label{fig:app_1}
\end{figure*}
\begin{table}
\centering
\caption{Total variation (TV) distance between the ID measure $\nu(x)$ and the predictive distribution $p_q(x)$ obtained from the optimization of the different free energy criteria.}
\begin{tabular}{@{}lllllll@{}}
\toprule
   &  $t=1 $ &  $t={0.9}$  & $t={0.7}$  & $t={0.5}$   & $t={0.3}$  & $t={0.1}$ \\ \midrule
$m=1$  & $0.59$               & $0.42$ & $0.27$ &$0.18$ &$\mathbf{0.16}$  &$0.18$ \\
$m=2$ & $0.44$               & $0.32$ & $0.22$ &$0.17$ &$\mathbf{0.15}$  &$0.15$ \\
$m=5$ & $0.34$               & $0.32$ & $0.23$ &$0.18$ &$0.15$  &$\mathbf{0.14}$\\
$m=10$  & $0.34$               & $0.30$ & $0.24$ &$0.19$ &$\mathbf{0.15}$  &$0.16$\\
\end{tabular}
\label{tab:first_appendix}
\end{table}

\subsection{Details on the Toy Example of Figure \ref{fig:toy_example_comparison}}
In the toy example of Figure \ref{fig:toy_example_comparison}, the ID distribution $\nu(x)$ is a two component Gaussian mixture with means $\{-2,2\}$, variance equal to 2, and mixing coefficients $\{0.3,0.7\}$, respectively. The OOD distribution $\xi(x)$ is modelled using a Gaussian distribution with mean -8 and variance equal to 1.

The probabilistic model is a Gaussian unit variance $p_{\theta}(x)=\mathcal{N}(x|\theta,1)$, the ensembling distribution $q(\theta)$ is represented by a discrete probability supported on 500 evenly spaced values in the interval $[-30,30]$, and the prior is $p(\theta)=\mathcal{N}(\theta|0,9)$. For a given $m$, $\beta$ and $t$, the optimized ensembling distribution is obtained applying the fixed-point iteration in Theorem \ref{thm1}, i.e.,
\begin{align}
    q^+(\theta)&=p(\theta)\exp\Bigg\{\beta \hspace{-0.2em}\sum_{\theta_1,\dots,\theta_{m-1}}\prod^{m-1}_{i=1}q^t(\theta_i)\cdot\nonumber\\
    &\hspace{5.2em}\cdot\log_t\Bigg(\hspace{-0.2em}\frac{\sum^{m-1}_{j=1}p(x|\theta_j)+p(x|\theta)}{m}\Bigg)\Bigg\},\\
    q^{t+1}(\theta)&=(1-\alpha)q^{t}(\theta)+\alpha\frac{q^+(\theta)}{\sum_\theta q^+(\theta)},
\end{align}
for $\alpha\in(0,1)$.

 In Figure \ref{fig:app_1} we report the optimized predictive distributions produced by the above procedure for $\beta=1$, $m=\{1,2,5,20\}$ and $t=\{1,0.9,0.7,0.5,0.3,0.1\}$. 
As $m$ grows larger, the multi-sample bound on the predictive risk becomes tighter. As a result, the predictive distribution becomes more expressive, and it covers all the data points. The use of generalized logarithms offers increased robustness against the outlier data point, and leads to predictive distributions that are more concentrated around the ID measure. In Table \ref{tab:first_appendix} we report the total variation distance between the ID measure and the predictive distribution $p_q(x)$. The proposed robust $m$-free energy criterion consistently outperforms the standard criterion by halving the total variation distance form the ID measure for $t=0.3$.

\subsection{Details and Further Results for the Classification Example in Sec. \ref{sec:class}}
\begin{figure}
    \centering
    \hspace{-1em}
    \includegraphics[width=0.8\columnwidth]{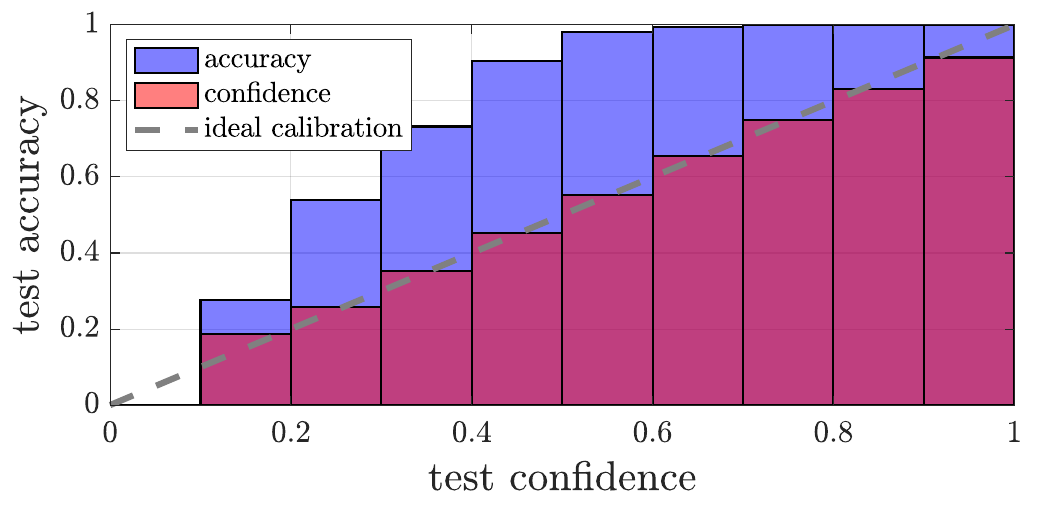}
    \caption{Reliability diagram of deep ensembles \cite{lakshminarayanan2017simple}.}
    \label{fig:class_acc_and_ece_reliability_deep}
\end{figure}

In Figure \ref{fig:class_acc_and_ece}, we used expected calibration error (ECE) \cite{guo2017calibration} to assess the quality of uncertainty quantification of the classifier. In this section, we formally define the ECE, along with the related visual tool of reliability diagrams \cite{degroot1983comparison}, and present additional results using reliability diagrams.

Consider a probabilistic parametric classifier $p(b|a,\theta)$, where $b \in \{ 1,\ldots,C\}$ represents the label and $a$ the covariate. The \emph{confidence} level assigned by the model to the predicted label
\begin{align}
    \hat{b}(a) = \argmax_{b} p(b|a,\theta)
\end{align} given the covariate $a$  is given as \cite{guo2017calibration}
\begin{align}
    \hat{p}(a) = \max_{b} p(b|a,\theta).
\end{align}
 \emph{Perfect calibration} corresponds to the equality \cite{guo2017calibration} 
\begin{align}
    \mathbb{P}( \hat{b}(a) = b | \hat{p}(a) = p ) = p, \text{  }\forall p \in [0,1],
    \label{eq:perfect_cali_app}
\end{align}
where the probability is taken over the ID sampling distribution $\nu(a,b)$. This equality expresses the condition that the probability of a correct decision for inputs with confidence level $p$ equals $p$ for all $p \in [0,1]$. In words, confidence equals accuracy.

The ECE and reliability diagram provide means to quantify the extent to which the perfect calibration condition (\ref{eq:perfect_cali_app}) is satisfied. To start, the probability interval $[0,1]$ is divided into $K$ bins, with the $k$-th bin being interval $(\frac{k-1}{K}, \frac{k}{K}]$. Assume that we have access to test data from the ID distribution. Denote as $\mathcal{B}_k$ the set of data points $(a,b)$ in such test set for which the confidence $\hat{p}(a)$ lies within the $k$-th bin, i.e., $\hat{p}(a) \in (\frac{k-1}{K}, \frac{k}{K}]$. The average accuracy of the  predictions for data points in $\mathcal{B}_k$ is defined as
\begin{align}
    \text{acc}(\mathcal{B}_k) = \frac{1}{|\mathcal{B}_k|} \sum_{a \in \mathcal{B}_k} \mathbf{1}(\hat{b}(a) = b),
\end{align}
with $\mathbf{1}(\cdot)$ being indicator function, $b$ being the label corresponding to $a$ in the given data point $(a,b)$, and $|\mathcal{B}_k|$ denoting the number of total samples in the $k$-th bin $\mathcal{B}_k$. Similarly, the average confidence of the predictions for covariates in $\mathcal{B}_k$ can be written as
\begin{align}
    \text{conf}(\mathcal{B}_k) = \frac{1}{|\mathcal{B}_k|} \sum_{a \in \mathcal{B}_k} \hat{p}(a).
\end{align}
Note that perfectly calibrated model $p(b|a,\theta)$ would have $\text{acc}(\mathcal{B}_k) = \text{conf}(\mathcal{B}_k)$ for all $k \in \{1,\ldots,K\}$ in the limit of a sufficiently large data set.

\begin{figure*}[t]
    \centering
    \hspace{-1em}
    \includegraphics[width=1\textwidth]{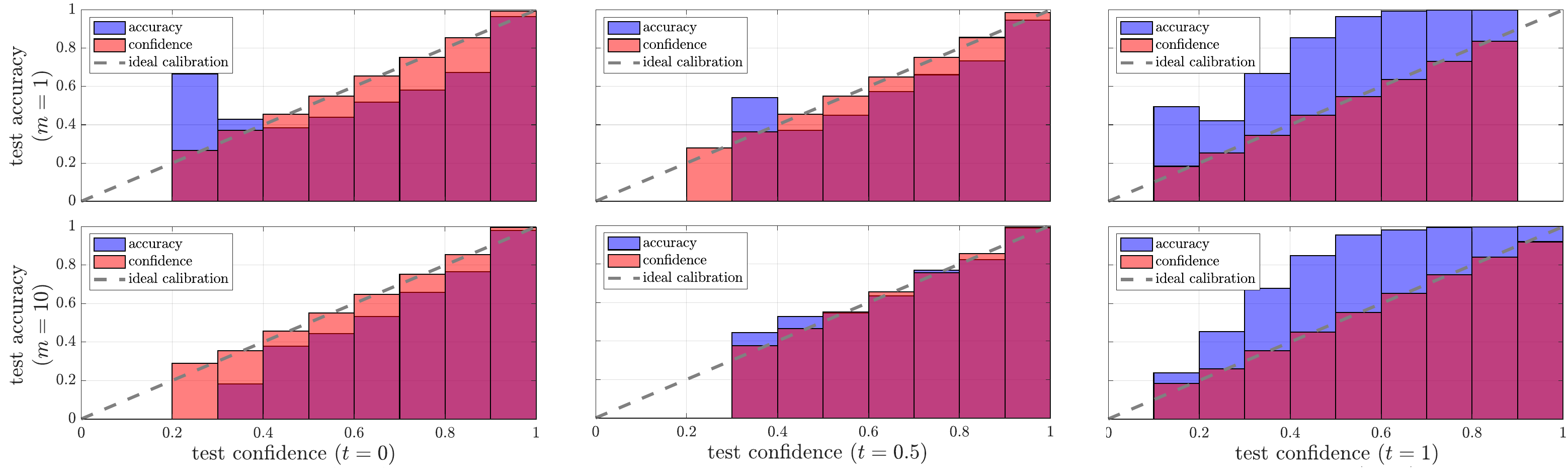}
    \caption{Reliability diagrams of robust Gibbs predictor that optimizes $\mathcal{J}_t^1$ (top); and proposed robust ensemble predictor that optimizes $\mathcal{J}_t^{10}$ (bottom) under contamination ratio $\epsilon=0.3$ for different $t=0,0.5,1$.     }
    \label{fig:class_acc_and_ece_reliability_proposed}
\end{figure*}
\subsubsection{Expected Calibration Error (ECE) \cite{guo2017calibration}} 
ECE quantifies the amount of miscalibration by computing the weighted average of the differences between accuracy and confidence levels across the bins, i.e.,
\begin{align}
    \text{ECE} = \sum_{k=1}^K \frac{|\mathcal{B}_k|}{\sum_{k=1}^K|\mathcal{B}_k| } \Big| \text{acc}(\mathcal{B}_k) - \text{conf}(\mathcal{B}_k) \Big|.
\end{align}

\subsubsection{Reliability Diagrams} 
Since the ECE quantifies uncertainty by taking an  average over the bins, it cannot provide insights into the individual calibration performance per bin. In contrast, reliability diagrams plot the accuracy $\text{acc}(\mathcal{B}_k)$ versus the confidence $\text{conf}(\mathcal{B}_k)$ as a function of the bin index $k$, hence offering a finer-grained understanding of the calibration of the predictor.

\subsubsection{Additional Results} For the MNIST image classification problem considered in Section \ref{sec:class}, Figure \ref{fig:class_acc_and_ece_reliability_deep} plots for reference the reliability diagrams for deep ensembles \cite{lakshminarayanan2017simple}, while Figure \ref{fig:class_acc_and_ece_reliability_proposed} reports reliability diagrams for the proposed classifiers with different values of $m$ and $t$. The figures illustrate that using the standard log-loss ($t=1$) tends to yield poorly calibrated decisions (Figure \ref{fig:class_acc_and_ece_reliability_deep} and Figure \ref{fig:class_acc_and_ece_reliability_proposed} (right)), while the proposed robust ensemble predictor can accurately quantify uncertainty using $t=0.5$ (Figure \ref{fig:class_acc_and_ece_reliability_proposed} (bottom, middle)). It is also noted that setting $t=1$ is seen to yield underconfident predictions due to the presence of outliers, while a  decrease in $t$ leads to overconfident decision due to the reduced expressiveness of $t$-logarithms. A proper choice of $t$ leads to well-calibrated, robust prediction.

\end{document}